\documentclass[preprint,12pt,authoryear]{elsarticle}
\usepackage{fix-cm}
\usepackage{amsmath,amsfonts}
\usepackage{algorithmic}
\usepackage{algorithm}
\usepackage{array}
\usepackage[caption=false,font=normalsize,labelfont=sf,textfont=sf]{subfig}
\usepackage{textcomp}
\usepackage{stfloats}
\usepackage{url}
\usepackage{verbatim}
\usepackage{graphicx}
\usepackage{cite}

\usepackage{amssymb}
\usepackage{mathtools}
\usepackage{amsthm}

\usepackage[utf8]{inputenc} 
\usepackage[T1]{fontenc}    
\usepackage{nicefrac}       
\usepackage{xcolor}         
\usepackage{float}
\usepackage{bookmark}
\usepackage{multirow}
\usepackage{makecell}
\usepackage{threeparttable}
\usepackage{tabularx}
\usepackage{bm}
\usepackage{booktabs}
\usepackage{natbib}

\newcolumntype{C}{>{\footnotesize\centering\arraybackslash}X}
\newcolumntype{P}[1]{>{\footnotesize\centering\arraybackslash}p{#1}}

\graphicspath{{./images/}}

\makeatletter

\newcommand{\mycorsymbol}[1]{%
  \ifcase#1\relax
  \or \ensuremath{*}
  \or \ensuremath{\dagger}
  \or \ensuremath{\ddagger}
  \else \ensuremath{\ast}%
  \fi
}

\renewcommand{\corref}[1]{%
  \edef\cnotenum{\elsRef{#1}}%
  \edef\@corref{%
    \noexpand\mycorsymbol{\cnotenum}%
    \noexpand\hskip-1pt
  }%
}

\renewcommand{\cortext}[2][]{%
  \g@addto@macro\@cornotes{%
    \refstepcounter{cnote}%
    \elsLabel{#1}%
    \def\thefootnote{\mycorsymbol{\value{cnote}}}%
    \footnotetext{#2}%
  }%
}

\makeatother

\journal{Chaos, Solitons and Fractals}

\begin{document}

\begin{frontmatter}
  \title{Ternary Spiking Neural Networks Enhanced by Complemented Neurons and Membrane Potential Aggregation}

  \author[1]{Boxuan Zhang\corref{cor1}}

  \author[1]{Jiaxin Wang\corref{cor1}}

  \author[2]{Zhen Xu}

  \author[1,3]{Kuan Tao\corref{cor2}}
  \ead{taokuan@bsu.edu.cn}

  \cortext[cor1]{Equal contribution}
  \cortext[cor2]{Corresponding author}

  \affiliation[1]{organization={School of Sports Engineering, Beijing Sport University},
      city={Beijing},
      postcode={100091},
      country={China}}

  \affiliation[2]{organization={School of Mathematics, Statistics and Mechanics, Beijing University of Technology},
      city={Beijing},
      postcode={100124},
      country={China}}
  \affiliation[3]{organization={Guangdong Key Laboratory of Human Sports Performance Science},
      city={Guangzhou},
      postcode={510500},
      state={Guangdong},
      country={China}}
  
  \begin{abstract}
    Spiking Neural Networks (SNNs) are promising energy-efficient models and powerful framworks of modeling neuron dynamics. However, existing binary spiking neurons exhibit limited biological plausibilities and low information capacity. Recently developed ternary spiking neuron possesses higher consistency with biological principles (e.g. excitation-inhibition balance mechanism). Despite of this, the ternary spiking neuron suffers from defects including iterative information loss, temporal gradient vanishing and irregular distributions of membrane potentials. To address these issues, we propose Complemented Ternary Spiking Neuron (CTSN), a ternary spiking neuron model that incorporates an learnable complemental term to store information from historical inputs. CTSN effectively improves the deficiencies of ternary spiking neuron, while the embedded learnable factors enable CTSN to adaptively adjust neuron dynamics, providing strong neural heterogeneity. Furthermore, based on the temporal evolution features of ternary spiking neurons' membrane potential distributions, we propose the Temporal Membrane Potential Regularization (TMPR) training method. TMPR introduces time-varying regularization strategy utilizing membrane potentials, furhter enhancing the training process by aggregating the membrane potential in each layer toward zero for better training stability. We validate our methods through extensive experiments across multiple datasets, demonstrating consistent performance improvements.
  \end{abstract}


\end{frontmatter}

\section{Introduction}
Spiking Neural Networks (SNNs) \citep{maass1997snn} have received widespread attention in recent years \citep{roy2019towards,schuman2022opportunities,kudithipudi2025neuromorphic}. Inspired by the biological neurons, SNNs adopt spiking neurons such as the Leaky Integrate-and-Fire (LIF) model \citep{gerstner2014lif}, realizing information transmission through binary spike-driven communication. This spike-driven communication equips SNNs multiplication-free characteristic and sparsity in information processing, offering SNNs excellent computational efficiency and low power consumption, making them powerful frameworks for modeling biological neural networks \citep{schmidt2018multi,siegle2021survey}. Spiking neuron models typically implement binary spiking mechanism, which allows neurons to fire spikes with values of either 0 or 1. This two-state characteristic of binary spiking mechanism constrains SNNs' information capacity. Moreover, binary spiking mechanism limits the potential of SNNs to simulate complex behaviors observed in biological neurons, such as excitation-inhibition balance mechanism \citep{xue2014equalizing}. Previous study \citep{guo2024ternary} proposed a ternary spiking neuron model that fires spikes in $\{-1,0,1\}$.
Ternary spiking neuron possesses higher consistency with biological principles (see Section. \ref{sec:biological}) and greater information capacity while retaining the multiplication-free and event-driven features, effectively elevates SNNs' performance. To gain a deeper understanding of this model, we perform in-depth analysis in Section. \ref{sec:motivation}, revealing its existing deficiencies including iterative information loss, temporal gradient vanishing and irregular distributions of membrane potentials. Besides, the current form of ternary spiking neuron neglects the heterogeneity of neuron dynamics \citep{yao2022glif,zheng2024dhsnn}, which is one of the primary factors leading to suboptimal performance of SNNs. In this case, further discussions and modifications are required to unleash the potential of this promising neuron model.

In this paper, utilizing the ternary spiking neuron as a baseline, we propose the Complemented Ternary Spiking Neuron (CTSN) model. In CTSN, we introduce a learnable complemental term $h(t)$ in the integration process, which can adaptively capture and maintain information from historical inputs, equiping neurons with stronger heterogeneity. CTSN continuously reinjects the information contained in previous inputs to the membrane potential as timestep increases, creates extra backpropagation paths in temporal gradient computation, and smooths the distribution of membrane potential.
\begin{figure}[!t]
  \centering
  \includegraphics[width=\textwidth]{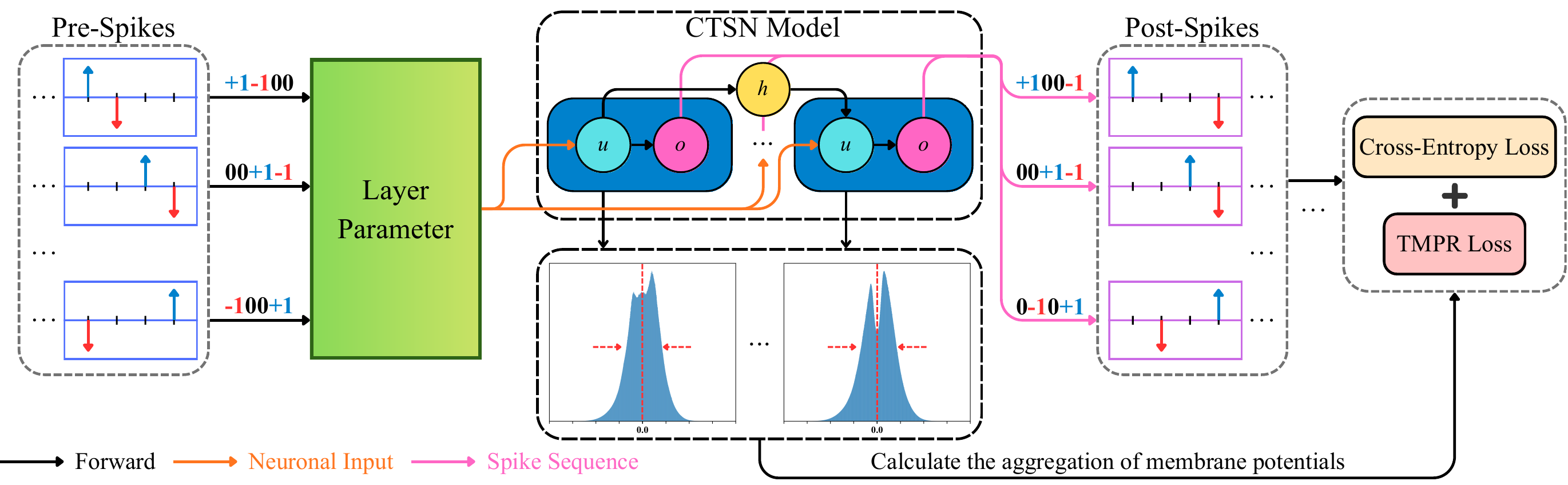}
  \captionsetup{width=\textwidth}
  \caption{The overall workflow of CTSN and TMPR.}
  \label{fig:workflow}
\end{figure}

By analyzing the temporal evolution of ternary spiking neuron's membrane potential distribution, we propose the Temporal Membrane Potential Regularization (TMPR) training method. TMPR introduces a spatio-temporal regularization strategy that exploits the aggregation degree of membrane potentials across multilayers and timesteps, further enhances the gradient flow in backpropagation stage. Under the joint promotion of CTSN and TMPR, the performance of ternary SNNs are effectively elevated. Figure. \ref{fig:workflow} depicts the workflow of our methods. Our main contributions are:
\begin{itemize}
    \item Based on the ternary spiking neuron, we propose the Complemented Ternary Spiking Neuron (CTSN) model with a learnable complemental integration term to preserve historical information to improve temporal gradient propagation, helps to mitigate iterative information loss and smooth irregular membrane potential distributions.
    \item We introduce the TMPR training method that exploits the aggregation degree of membrane potentials across layers and timesteps as additional regularization terms, enhancing gradient flow of ternary SNNs during training.
    \item We validate the effectiveness of our methods through extensive experiments, and the results show that our methods consistently outperform existing SNN methods across multiple tasks.
\end{itemize}

\section{Related Work}
\textbf{Variants of Spiking Neuron Models.} Multiple variants of spiking neuron models are proposed in previous studies and these models can be divided into two categories: binary and multi-nary. Various binary neuron models with enhanced neuron dynamics are developed in prior works \citep{fang2021plif,yao2022glif,wang2022ltmd,huang2024clif}. In parallel, binary spiking neurons with bio-inspired structures such as multi-branch \citep{zheng2024dhsnn} and multi-compartment \citep{capone2023beyond,shibo2025tslif} have been proposed for modeling specific tasks (e.g. long sequence modeling). Despite their high biological plausibility and outstanding performance on their designed tasks, these models not only introduce significant computational complexity but also cannot be generalized to various network backbones and general scenarios. To increase the representation space of spiking neurons, recent works propose neuron models with multi-nary spiking mechanism for extending the binary spike space $\{0,1\}$ to a wider range. \citep{sun2022deep} proposed a neuron model that emits $\{0,1,2\}$ spikes. This model increases the information bandwidth of spike sequence, but fails to enjoy multiplication-free characteristic, resulting in lower energy efficiency. \citep{wang2025mmdend} proposed a multi-branch multi-compartment neuron model with the integer set as its spike space. However, this model is specifically designed for long sequence modeling, with poor generalizability and high computational complexity. \citep{guo2024ternary} developed the ternary spiking neuron model for image classification tasks. 
However, the improvement to SNNs by merely expanding spike spaces is limited. In this paper, our CTSN incorporates bio-inspired neuron behaviors and $\{-1,0,1\}$ ternary spiking mechanism to advance the performance of SNNs without adding significant complexity to the neuron.

\textbf{Gradient-based Learning of SNNs.} Gradient-based learning of SNNs follows the idea of backpropagation in ANN \citep{rumelhart1986bp}, which essentially treats SNNs as RNNs and directly trains them via surrogate gradients \citep{neftci2019surrogate} and backpropagation through time (BPTT) \citep{wu2018bptt} frameworks. During the forward stage, spikes are emitted by the non-differentiable spiking function, while the non-differentiable terms are generally replaced by surrogate gradients in backpropagation. In order to further improve the efficiency of direct training, innovative approaches \citep{wu2019neunrom,fang2021sew,deng2022tet,meng2023sltt,shen2024implts,liu2025deeptage,ding2025rethinking,zhang2026trt} are proposed to accelerate the convergence of SNNs. Besides, some studies \citep{mukhoty2023localzo,wang2023asgl} focus on surrogate gradient to improve the accuracy of SNNs. However, few works link the training process directly with neuron dynamics. \citep{zheng2021tdbn} proposed tdBN, which combines neuron threshold with batch normalization for efficient training. \citep{guo2023rmp} utilized membrane potentials as regularization terms to reduce the quantization error of SNNs. Still, these works neglect the inherent temporal characteristics of neuron dynamics. In this paper, we construct time-varying regularization strategy aligned with the temporal evolution of membrane potential distributions to constrain membrane potentials and mitigate temporal gradient vanishing.

\begin{figure}[!t]
  \centering
  \includegraphics[width=\textwidth]{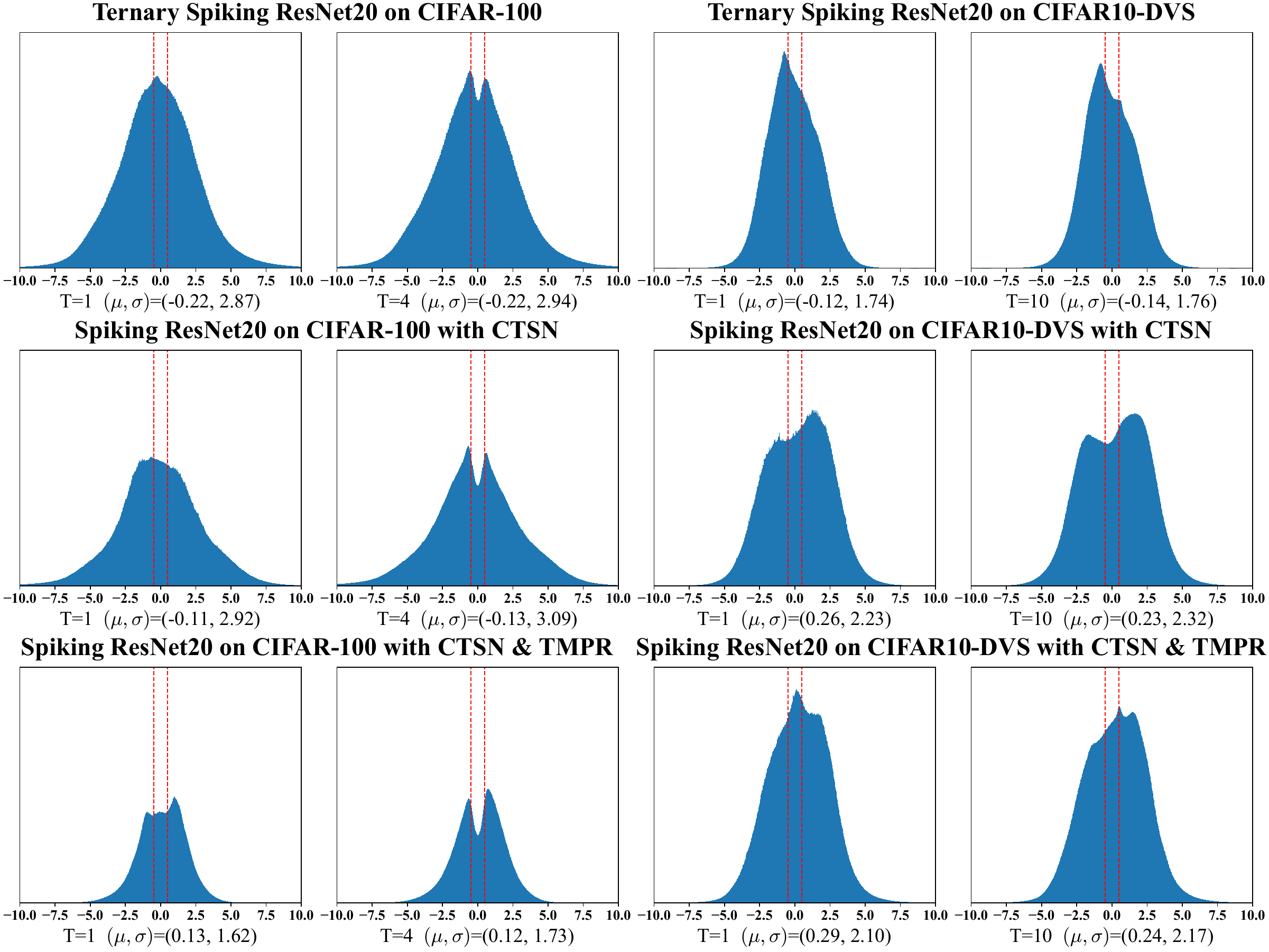}
  \captionsetup{width=\textwidth}
  \caption{Membrane potential distributions of the first and the last timestep. Upper left: Ternary spiking ResNet20 on CIFAR-100 (4 timesteps). Upper right: Ternary spiking ResNet20 on CIFAR10-DVS (10 timesteps). Middle left: Spiking ResNet20 on CIFAR-100 (4 timesteps) with CTSN. Middle right: Spiking ResNet20 on CIFAR10-DVS (10 timesteps) with CTSN. Lower left: Spiking ResNet20 on CIFAR100 (4 timesteps) with CTSN and TMPR. Lower right: Spiking ResNet20 on CIFAR10-DVS (10 timesteps) with CTSN and TMPR. See \ref{sec:appendix_membrane_visulization} for detailed visualizations.}
  \label{fig:membrane_potential_distribution}
\end{figure}
\section{Preliminary}
We adopt the ternary spiking neuron model proposed in \citep{guo2024ternary} as the baseline ternary neuron model due to its event-driven and multiplication-free features, which is described as
\begin{equation}
  u(t)=\tau u(t-1)(1-|o(t-1)|)+x(t),
  \label{eq:ternary_u}
\end{equation}
\begin{equation}
  x(t)=\sum_{l=1}^{L}\mathbf{W}^lo^{l-1}(t)+b^l,
\end{equation}
and
\begin{equation}
  \begin{aligned}
    o(t) & = \Theta(u(t),V_{th})\\
    & =
    \begin{cases}
      1,  & \text{if }u(t)\ge V_{th} \\
      -1, & \text{if }u(t)\le -V_{th} \\
      0,  & otherwise                \\
    \end{cases} \\
  \end{aligned},
  \label{eq:ternary_fire}
\end{equation}
where $l$ is is the layer indexis the layer index, $V_{th}$ is the firing threshold, and $\tau$ is the time constant. The membrane potential $u(t)$ is updated at each timestep $t$ based on the previous membrane potential $u(t-1)$ and input $x(t)$, which is producted by the $l^{th}$ weight matrix $\mathbf{W}^l$, bias term $b^l$, and spikes of the previous layer neruon $o^{l-1}(t)$. The neuron fires when $u(t)\ge V_{th}$ or $u(t)\le -V_{th}$, and the membrane potential will be reset to $0$ after firing, which is known as the hard reset. 

The network's final output at $t^{th}$ timestep is $O(t)=\phi(\mathbf{X}(t))$, where $\phi$ denotes the network and $\mathbf{X}$ is the external input sequence. The prediction is based on the average output over all timesteps $\frac{1}{T}\sum_{t=1}^{T}O(t)$, where $T>0$. The loss function is defined as
\begin{equation}
  \mathcal{L}=\ell\biggl(\frac{1}{T}\sum_{t=1}^{T}O(t),\mathbf{Y}\biggr),
\end{equation}
where $\mathbf{Y}$ is the label, and $\ell$ represents the cross-entropy loss function.

Since the spike activity is not differentiable, we use a rectangular-shaped surrogate gradient function to calculate the spike derivative:
\begin{equation}
  \frac{\partial o(t)}{\partial u(t)} \approx \mathbb{H}\big(u(t)\big) = \text{Sign}(|u(t)| - V_{th} < a),
  \label{eq:rectangular}
\end{equation}
where $\text{Sign}(\cdot)$ denotes the sign function and $a$ is a hyperparameter controlling the width. In this work, we set $a=V_{th}=0.5$.

\section{Motivation}
\label{sec:motivation}
This section identifies three limitations of the ternary spiking neuron that motivate our approach: iterative information loss caused by hard resets, temporal gradient vanishing during backpropagation, and irregular membrane potential distributions that may result in either excessive activation or neuronal silence. The following subsections examine these issues in turn.

\subsection{Iterative Information Loss of Ternary Spiking Neuron}
\label{sec:motivation_information_loss}
SNN can be seen as a special case of RNN, where the hidden state is the membrane potential $u(t)$. From this perspective, inspired by the iterative firing process of binary spiking neurons \citep{huang2025arlif}, the ternary spiking neuron model depicted by Eq. \ref{eq:ternary_u} - \ref{eq:ternary_fire} can be derived into
\begin{equation}
  \begin{split}
    o(t)&=\Theta\big(u(t), V_{th}\big)\\
    &=\Theta\big(\tau u(t-1)(1-|o(t-1)|)+x(t), V_{th}\big)\\
    &=\Theta\biggl(\tau (1-|o(t-1)|)\big(\tau (1-|o(t-2)|)u(t-2)\\
    &\quad +x(t-1)\big)+x(t), V_{th}\biggr)\\
    &\quad \dots\\
    &=\Theta\biggl(x(t)+\sum_{i=1}^{t-1}\biggl(\tau^{t-i}x(i)\prod_{j=i}^{t-1}\bigl(1-|o(j)|\bigr)\biggr), V_{th}\biggr),
  \end{split}
  \label{eq:iterative}
\end{equation}
where $\Theta(\cdot,V_{th})$ denotes the firing process described in Eq. \ref{eq:ternary_fire}. From Eq. \ref{eq:iterative}, it saliently yields that firing status of a ternary spiking neuron is highly related to the inputs and spikes in previous timesteps. Due to the hard reset mechanism, the information stored in previous inputs is lost after each firing. Therefore, similar to binary spiking neurons with hard reset, the ternary spiking neuron model is a memoryless model. Once the neuron fires a spike at a given timestep, it will be unable to utilize any information from the inputs arrived in previous timesteps. For binary spiking neurons, the soft reset mechanism \citep{guo2022reducing} is generally implemented to address this issue. We conducted experiments to test the soft reset mechanism on ternary spiking neuron. In this case, the membrane potential update process is modified as
\begin{equation}
  u(t)=\tau \big(u(t-1)-o(t-1)V_{th}\big)+x(t).
\end{equation}
However, the accuracy results turn out to be unsatisfactory (results and detailed analysis are in Section. \ref{sec:ablation_studies}). This indicates that directly applying the standard soft reset mechanism to ternary spiking neurons is not effective. 

\subsection{Temporal Gradient Vanishing in Ternary Spiking Neuron}
\label{sec:motivation_gradient}
Ternary SNNs are trained through BPTT \citep{wu2018bptt}. Similar to binary SNNs \citep{meng2023sltt,huang2024clif}, the backpropagation process of gradients involves the temporal dimension, which can be described as
\begin{equation}
  \begin{split}
    \frac{\partial \mathcal{L}}{\partial \mathbf{W}^l}&=\sum_{t=1}^{T}\frac{\partial \mathcal{L}}{\partial u^l(t)}\frac{\partial u^l(t)}{\partial \mathbf{W}^l}\\
    &=\sum_{t=1}^{T}\biggl(\nabla_{u^l}\mathcal{P}(t)+\nabla_{u^l}\mathcal{T}(t)\biggr)\frac{\partial u^l(t)}{\partial \mathbf{W}^l}\\
    &=\sum_{t=1}^{T}\biggl(\underbrace{\frac{\partial \mathcal{L}}{\partial o^l(t)}\frac{\partial o^l(t)}{\partial u^l(t)}}_{\nabla_{u^l}\mathcal{P}(t)}+\underbrace{\sum_{t'=t+1}^{T}\frac{\partial \mathcal{L}}{\partial o^l(t')}\frac{\partial o^l(t')}{\partial u^l(t')}\prod_{t''=1}^{t'-t}\epsilon^l(t'-t'')}_{\nabla_{u^l}\mathcal{T}(t)}\biggr)\frac{\partial u^l(t)}{\partial \mathbf{W}^l},
  \end{split}
  \label{eq:gradient}
\end{equation}
where $l=L,L-1,\dots,1$, $t'\in[t+1,T]$. For different layers, $\frac{\partial \mathcal{L}}{\partial o^l(t)}$ term is expressed in different forms, which is calculated as
\begin{equation}
  \frac{\partial \mathcal{L}}{\partial o^l(t)}=
  \begin{cases}
    \frac{\partial \mathcal{L}}{\partial o^{L}(t)}, & l=L \\
    \frac{\partial \mathcal{L}}{\partial u^{l+1}(t)}\frac{\partial u^{l+1}(t)}{\partial o^l(t)}, & L=L-1,L-2 \dots 1 \\
  \end{cases}.
  \label{eq:gradient_layers}
\end{equation}
$\epsilon^l(t)$ is defined as
\begin{equation}
    \begin{split}
        \epsilon^l(t) & =\frac{\partial u^l(t+1)}{\partial u^l(t)}+\frac{\partial u^l(t+1)}{\partial |o^l(t)|}\frac{\partial |o^l(t)|}{\partial u^l(t)}\\
                 & =\tau\big(1-|o^l(t)|-\text{Sign}(o^l(t))u^l(t)\mathbb{H}[u^l(t)]\big).
    \end{split}
    \label{eq:epsilon}
\end{equation}
In this case, we obtain
\begin{equation}
  \kappa^l(t)=1-|o^l(t)|-\text{Sign}(o^l(t))u^l(t)\mathbb{H}[u^l(t)],
\end{equation}
$\nabla_{\mathbf{u}^l}\mathcal{T}(t)$ can be rewritten as
\begin{equation}
  \begin{split}
    \nabla_{\mathbf{u}^l}\mathcal{T}(t)&=\sum_{t'=t+1}^{T}\frac{\partial \mathcal{L}}{\partial o^l(t')}\frac{\partial o^l(t')}{\partial u^l(t')}\underbrace{\tau^{t'-t}}_{\text{Part I}}\prod_{t''=1}^{t'-t}\underbrace{\kappa^l(t'-t'')}_{\text{Part II}},
  \end{split}
  \label{eq:temporal_gradient_ternary}
\end{equation}
where term $\mathbb{H}(\cdot)$ denotes the rectangle-shaped surrogate gradient we used in this work. The detailed derivations are given in \ref{sec:appendix_gradient_ternary}. For \textbf{Part I}, since the time constant $\tau$ is usually smaller than $1$ (e.g. $\tau=0.5$ \citep{du2025flexibility} and $\tau=0.25$ \citep{zheng2021going}), $\tau^{t'-t}$ eventually converges to $0$ as $(t'-t)$ increases. Therefore, the diagonal values of matrix $\prod_{t''=1}^{t'-t}\epsilon^l(t'-t'')$ are smaller than the single term $\epsilon^l(t'-t'')$, causing the vanishment of temporal gradients in early timesteps \citep{meng2023sltt}. \textbf{Part II} can be expressed as
\begin{equation}
  \begin{split}
    \kappa^l(t)
    =\begin{cases}
        u^l(t), & -\frac{3}{2}V_{th}\le u^l(t)<-V_{th} \\
        (1+u^l(t)), & -V_{th}\le u^l(t)<0 \\
        (1-u^l(t)), & 0<u^l(t)\le V_{th} \\
        (-1+u^l(t)), & V_{th}<u^l(t)\le \frac{3}{2}V_{th} \\
        0,  & otherwise \\
      \end{cases}.
  \end{split}
  \label{eq:part-II}
\end{equation}
From Eq. \ref{eq:part-II}, it is deduced that $\kappa^l(t)=0$ always holds when $|u(t)|\ge \frac{3}{2}V_{th}$, even if the neuron actually fires a spike. In other words, within the temporal range of $(t+1,T)$, as long as the neuron fires at least once under the condition of $|u(t)|\ge \frac{3}{2}V_{th}$, the corresponding temporal gradient will vanish. This is an unavoidable issue in the ternary spiking neuron.

\subsection{Irregular Membrane Potential Distribution of Ternary Spiking Neuron}
As can be seen from Figure. \ref{fig:membrane_potential_distribution}, the membrane potential distributions of the ternary spiking neuron exhibits distinct characteristics across different types of data. This phenomenon is most likely attributed to the inherent sparsity of neuromorphic data. On static image data, the membrane potential distribution exhibit irregular and threshold-deviated bimodal pattern as the timestep increases. This may lead to over activation, resulting in gradient vanishing as described in Eq. \ref{eq:part-II}. In contrast, for neuromorphic data, the membrane potential at later timesteps becomes more sharply peaked but does not display a distinct bimodal distribution. This may result in a large number of neurons remaining silent, thereby degrading the representational power of neurons. Detailed visualizations are provided in \ref{sec:appendix_membrane_visulization}. Therefore, tailored approaches are required to address the irregular membrane potential distribution of ternary spiking neurons on different data types.

\section{Methodology}
\subsection{Complemented Ternary Spiking Neuron}
\label{sec:ctsn}
Inspired by biological principles (see Section. \ref{sec:biological}), 
we propose to add an additional term to the ternary spiking neuron model.
The complemented ternary spiking neuron (CTSN) can be described by
\begin{equation}
  u(t)=\tau \tilde{u}(t-1)(1-|o(t-1)|),
  \label{eq:ctsn_u}
\end{equation}
\begin{equation}
  h(t)=\mathbb{G}\big(h(t-1),u(t),\alpha,\beta,\gamma\big),
\end{equation}
\begin{equation}
  \tilde{u}(t)=h(t)+x(t),
\end{equation}
and
\begin{equation}
  o(t) = \Theta(\tilde{u}(t),V_{th}).
  \label{eq:ctsn_fire}
\end{equation}
Term $\tilde{u}(t)$ denotes the membrane potential after integration, and $h(t)$ is the additional term. The initial value of $\tilde{u}(t)$ and $h(t)$ are $0$. Function $\mathbb{G}\big(h(t-1),u(t),\alpha,\beta,\gamma\big)$ defines the update process of the complemental membrane potential. For static image data, $\mathbb{G}(h(t-1),u(t),\alpha,\beta,\gamma)$ is defined as
\begin{equation}
  \begin{aligned}
    \mathbb{G} =
    \begin{cases}
      \alpha h(t-1) + \gamma u(t), & h(t-1)\ge 0 \\
      \beta h(t-1) + \gamma u(t), & h(t-1)<0 \\
    \end{cases} \\
  \end{aligned}.
  \label{eq:g_static}
\end{equation}
For neuromorphic data, $\mathbb{G}(h(t-1),u(t),\alpha,\beta,\gamma)$ is defined as
\begin{equation}
  \begin{aligned}
    \mathbb{G} =
    \begin{cases}
      \alpha h(t-1) + \beta u(t), & u(t)\ge 0 \\
      \alpha h(t-1) + \gamma u(t), & u(t)<0 \\
    \end{cases} \\
  \end{aligned}.
  \label{eq:g_neuromorphic}
\end{equation}
$\alpha, \beta, \gamma$ are learnable parameters. To ensure all these parameters are constrained in $(0, 1)$, we train parameters $\omega_\alpha, \omega_\beta, \omega_\gamma$ and set
\begin{equation}
  \alpha = \text{Sigmoid}(\omega_\alpha),
  \beta = \text{Sigmoid}(\omega_\beta),
  \gamma = \text{Sigmoid}(\omega_\gamma),
\end{equation}
with $\omega_\alpha, \omega_\beta, \omega_\gamma$ setting to $0$ by default. In this case, the initial values of $\alpha, \beta, \gamma$ are $0.5$.

The additional term $h(t)$ can be seen as a memristor without any reset mechanism. For those neurons that intermittently emit spikes, $h(t)$ can effectively retain information from inputs arrived in previous timesteps, and reinject these information to the neuron in later timesteps, thereby partially addressing the iterative information loss problem discussed in Section. \ref{sec:motivation_information_loss}. Moreover, $h(t)$ also smooths the membrane potential and contributes the temporal gradient during backpropagation. The pseudo-code for the CTSN model is provided in Algorithm. \ref{alg:ctsn}.
\begin{algorithm}[!t]
    \caption{Core Function of CTSN Model.}
    \label{alg:ctsn}
    \begin{algorithmic}
        \STATE{\bfseries Input:} Data $\mathbf{x}$, total timesteps $T$, time constant $\tau$, threshold $V_{th}$, learnable parameters $\alpha,\beta,\gamma$, data type.
        \STATE{\bfseries Output:} Spike sequence $\mathbf{\mathcal{S}}$ and membrane potential train $\mathbf{\mathcal{U}}$.
        \STATE{\bfseries Initial:} $\mathbf{u}(t)=0$, $\mathbf{h}(t)=0$, $\mathbf{\tilde{u}}(t)=0$, $\mathbf{\mathcal{S}}=\{\}, \mathbf{\mathcal{U}}=\{\}$.
        \FOR{$t=1,\dots,T$}
            \STATE $\mathbf{u}(t)=\tau \mathbf{\tilde{u}}(t)$;
            \IF{\textup{processing static data}}
                \STATE $\mathbf{h}(t)=\alpha \text{ReLU}(\mathbf{h}(t-1))+\beta(-\text{ReLU}(-\mathbf{h}(t-1)))+\gamma \mathbf{u}(t)$;
            \ENDIF
            \IF{\textup{processing neuromorphic data}}
                \STATE $\mathbf{h}(t)=\alpha \mathbf{h}(t-1)+\beta \text{ReLU}(\mathbf{u}(t))+\gamma (-\text{ReLU}(-\mathbf{u}(t)))$;
            \ENDIF
        \STATE $\mathbf{\tilde{u}}(t)=\mathbf{h}(t)+\mathbf{x}(t)$;
        \STATE $\mathbf{\mathcal{U}}\cup\mathbf{\tilde{u}}(t)$;
        \STATE $\mathbf{o}(t)=\Theta(\mathbf{\tilde{u}}(t),V_{th})$;
        \STATE $\mathbf{\mathcal{S}}\cup\mathbf{o}(t)$;
        \STATE $\mathbf{\tilde{u}}(t)=\mathbf{\tilde{u}}(t)(1-|\mathbf{o}(t)|)$;
        \ENDFOR
        \STATE{\bfseries Return:} $\mathbf{\mathcal{S}}$, $\mathbf{\mathcal{U}}$.
    \end{algorithmic}
\end{algorithm}

\subsection{Temporal Membrane Potential Regularization}
As observed from the Figure. \ref{fig:membrane_potential_distribution}, the membrane potential of ternary spiking neurons inevitably differentiate into irregular distributions (e.g. bimodal distribution) as the timestep increases. In this case, we utilize the aggregation degree of the membrane potential as a regular term to enhance the training process, which is called Temporal Membrane Potential Regularization (TMPR) training method. TMPR is described as
\begin{equation}
  \mathcal{L} = \mathcal{L}_\mathrm{CE} + \mathcal{L}_\mathrm{TMPR},
  \label{eq:tmpr_total_loss}
\end{equation}
\begin{equation}
  \mathcal{L}_\mathrm{TMPR} = \frac{1}{TL} \sum_{t=1}^{T} \frac{\lambda}{t} \sum_{l=1}^{L} \frac{\tilde{u}^l(t)^2}{BD^l},
  \label{eq:tmpr}
\end{equation}
where $B$ denotes the batch size, $D^l$ denotes the length of the $l^{th}$ layer's neuron membrane potential $\tilde{u}^l(t)$ (for image data which in the shape of $[C,H,W]$, we can flatten the membrane potential so that it becomes a vector, in this case $D^l=C^l\times H^l\times W^l$), $T$ denotes the total timesteps, $L$ denotes the total number of layers, $\lambda$ is a regularization coefficient.

TMPR utilizes $\tilde{u}(t)^2$ to measure the aggregation degree of membrane potential. The strength of regularization is jointly controlled by the time step $t$ and $\lambda$. For early timesteps, it is necessary to impose strong regularization to force the membrane potential gathering around 0. Since the membrane potential will inevitably differentiate into irregular distribution in later time steps, we propose to reduce the strength of regularization to avoid the unstable training process caused by excessive penalties. This mechanism aligns with the temporal evolution trend of membrane potential distribution. The pseudo-code for the TMPR training method is provided in Algorithm. \ref{alg:tmpr}.
\begin{algorithm}[!t]
    \caption{Training an SNN with TMPR Method.}
    \label{alg:tmpr}
    \begin{algorithmic}
        \STATE{\bfseries Input:} SNN model, total timesteps $T$, training data, total number of training epochs $N$, number of training batches in one epoch $B$, layer number of the SNN model $L$, regularization coefficient $\lambda$.
        \STATE{\bfseries Output:} Trained SNN model.
        \FOR{$n=1,\dots,N$}
            \FOR{$i=1,\dots,B$}
                \STATE Get training data, and class label: $\mathbf{Y}^i$;
                \STATE Compute the SNN output $\mathbf{O}^i$;
                \STATE Compute classification loss $\mathcal{L}_\mathrm{CE}=\mathcal{L}_\mathrm{CE}(\mathbf{O}^i,\mathbf{Y}^i)$;
                \STATE Initialize $\mathbf{U}^i$ to $0$;
                \FOR{$l=1,\dots,L$}
                    \STATE Get and store each layer's flatten membrane potential of each timestep into $\mathbf{U}^i$;
                \ENDFOR
                \STATE Compute TMPR loss $\mathcal{L}_\mathrm{TMPR}$ based on Eq. \ref{eq:tmpr} using $\mathbf{U}^i$ and $\lambda$;
                \STATE Compute the total loss $\mathcal{L}=\mathcal{L}_\mathrm{CE}+\mathcal{L}_\mathrm{TMPR}$;
                \STATE Backpropagation and update model parameters;
            \ENDFOR
        \ENDFOR
        \STATE{\bfseries Return:} The trained SNN model.
    \end{algorithmic}
\end{algorithm}

\subsection{Gradient Analysis of CTSN and TMPR}
\label{sec:dynamic_and_theoretical_analysis}
With the implementation of CTSN and TMPR, the backpropagation process of gradients can be calculated as
\begin{equation}
  \frac{\partial \mathcal{L}}{\partial \mathbf{W}^l}=\sum_{t=1}^{T}\biggl(\underbrace{\frac{\partial \mathcal{L}_\mathrm{CE}}{\partial \tilde{u}^l(t)}}_{\text{Part I}}+\underbrace{\frac{\partial \mathcal{L}_\mathrm{TMPR}}{\partial \tilde{u}^l(t)}}_{\text{Part II}}\biggr)\frac{\partial \tilde{u}^l(t)}{\partial \mathbf{W}^l}.
\end{equation}
From Eq. \ref{eq:tmpr}, we can calculated \textbf{Part I} as
\begin{equation}
  \frac{\partial \mathcal{L}_\mathrm{CE}}{\partial \tilde{u}^l(t)}=\nabla_{\tilde{u}^l}\mathcal{P}(t)+\nabla_{\tilde{u}^l}\mathcal{T}(t).
  \label{eq:gradient_analysis_part_i}
\end{equation}
The $\nabla_{\tilde{u}^l}\mathcal{P}(t)$ shares the same form with the counterpart in ternary spiking neuron in Eq. \ref{eq:gradient}. The $\nabla_{\tilde{u}^l}\mathcal{T}(t)$ can be expanded as
\begin{equation}
  \begin{split}
    \nabla_{\tilde{u}^l}\mathcal{T}(t)&=\frac{\partial \mathcal{L}}{\partial o^l(t+1)}\frac{\partial o^l(t+1)}{\partial \tilde{u}^l(t+1)}\xi^l(t)\\
    &\quad + \sum_{t'=t+2}^{T}\frac{\partial \mathcal{L}}{\partial o^l(t')}\frac{\partial o^l(t')}{\partial \tilde{u}^l(t')}\prod_{t''=1}^{t'-t}\biggl(\xi^l(t'-t'')+\nabla_{h^l}\mathbb{G}^l\biggr).
  \end{split}
  \label{eq:temporal_gradient_ctsn}
\end{equation}
where $\nabla_{h^l}\mathbb{G}^l$ denotes the temporal gradient $\frac{\partial h(t+1)}{\partial h(t)}$, and $\xi^l(t)$ is defined as
\begin{equation}
  \begin{split}
    \xi^l(t) &= \frac{\partial h^l(t+1)}{\partial \tilde{u}^l(t)}+\frac{\partial h^l(t+1)}{\partial |o(t)|}\frac{\partial |o(t)|}{\partial \tilde{u}^l(t)}\\
    &= \nabla_{\tilde{u}^l}\mathbb{G}^l+\frac{\partial h^l(t+1)}{\partial |o(t)|}\frac{\partial |o(t)|}{\partial \tilde{u}^l(t)},
  \end{split}
  \label{eq:xi}
\end{equation}
where $\nabla_{\tilde{u}^l}\mathbb{G}^l$ denotes the spatial gradient $\frac{\partial h(t+1)}{\partial \tilde{u}^l(t)}$. Details of the above derivations are given in \ref{sec:appendix_gradient_ctsn}. As shown in Eq. \ref{eq:temporal_gradient_ctsn}, the backpropagation of CTSN degenerates into the same procedure as that of ternary spiking neuron when $T\le2$. When $T>2$, the unreset term $h(t)$ generates additional backpropagation path, magnifying the temporal gradient on intermediate timesteps to against the temporal gradient vanishing induced by the product operations. We believe this additional part contributes to the advancment on performance.

\textbf{Part II} can be calculated as
\begin{equation}
  \frac{\partial \mathcal{L}_\mathrm{TMPR}}{\partial \tilde{u}^l(t)}=\frac{2\lambda}{tTLB{D^l}}\tilde{u}^l(t).
  \label{eq:gradient_analysis_part_ii}
\end{equation}
As can be seen from Eq. \ref{eq:gradient_analysis_part_ii}, the TMPR method introduces direct backpropagation paths to the gradient of membrane potential on each timestep and layer within the SNN, which can be viewed as residual connections with varying strength over time on both spatial and temporal dimensions, pushing the parameters in each layer toward zero on early timesteps (i.e. smaller $t$).

CTSN and TMPR jointly enhance the backpropagation of gradients in both spatial and temporal dimensions, effectively mitigating the temporal gradient vanishing problem discussed in Section. \ref{sec:motivation_gradient}. This theoretical analysis aligns with our experimental observations in Section. \ref{sec:experimental_results}, where the integration of CTSN and TMPR leads to remarkable performance improvements in ternary spiking neural networks.

\section{Experimental Results}
\label{sec:experimental_results}
In this section, we validate the effectiveness of our proposed methods through a series of ablation studies. We perform a thorough analysis and comparison of our approach against other state-of-the-art methods using both static and neuromorphic datasets. We choose the commonly used backbones (including ResNet \citep{he2016resnet}, SEW-ResNet \citep{fang2021sew}, VGG16 \citep{simonyan2015vgg} and VGGSNN \citep{deng2022tet}) for each dataset to ensure a fair and comprehensive comparison with existing work.

\subsection{Ablation Studies}
\label{sec:ablation_studies}
\begin{table}[!t]
  \centering
  \caption{Comparative accuracy of ternary spiking ResNet20 with different reset mechanisms (\%).}
  \label{tab:soft_reset}
  \begin{threeparttable}
    \begin{tabularx}{0.9\columnwidth}{
    >{\hsize=1.7\hsize}C
    >{\hsize=0.3\hsize}C
    CC}
      \toprule
      & & \multicolumn{2}{c}{\textbf{Reset Mechanism}}\\
      \cmidrule(l){3-4}
      \multirow{-2.5}{*}{\textbf{Dataset}} & \multirow{-2.5}{*}{\textbf{T}} & Hard & Soft \\
      \toprule
      CIFAR-100 & 4 & $\mathbf{73.95}$ & $73.56$ \\
      \midrule
      CIFAR10-DVS & 10 & $\mathbf{80.30}$ & $80.20$\\
      \bottomrule
    \end{tabularx}
  \end{threeparttable}
\end{table}
\textbf{Comparison of Different Reset Mechanisms.} To explore the standard soft reset mechanism on baseline ternary spiking neurons, we conduct experiments on CIFAR-100 and CIFAR10-DVS datasets with ResNet20 as the backbone. The results in Table. \ref{tab:soft_reset} show that the soft-resetting ternary spiking neuron performs worse than the hard-resetting version on both datasets, indicating that directly applying the standard soft reset mechanism to ternary spiking neurons is not effective, specially designed reset mechanisms are required to address the issue of information loss and improve the performance of ternary SNNs. We perform detailed analysis on this phenomenon in \ref{sec:appendix_soft_reset}.

\begin{figure}[!t]
  \centering
  \includegraphics[width=0.6\columnwidth]{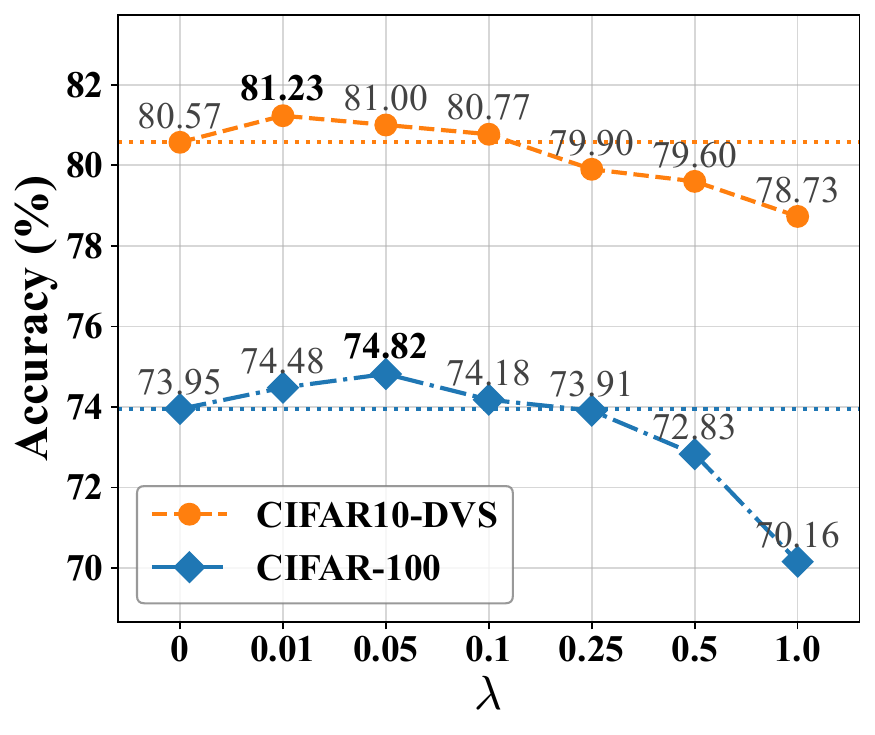}
  \captionsetup{width=\columnwidth}
  \caption{Influence of the hyperparameter $\lambda$ on performance.}
  \label{fig:lambda}
\end{figure}
\textbf{Effectiveness of CTSN and TMPR.} In order to verify the effectiveness of our CTSN and TMPR method, we conduct experiments on CIFAR-100 and CIFAR10-DVS datasets with ResNet20 as the backbone. As listed in Table. \ref{tab:ablation_of_effectiveness}, the networks solely implemented with CTSN or TMPR outperform those using ternary spiking neurons across different timestep settings on various datasets. Under the joint equipment of these two techniques, the classification accuracy of ResNet20 model raised from $80.57\%$ to $81.23\%$ on CIFAR10-DVS, $74.48\%$ to $75.12\%$ on CIFAR-100 when time length $T=6$, demonstrating the effectiveness of this method. We further conducted experiments with deeper networks, confirming that CTSN and TMPR remain effective across architectures of varying complexity (see \ref{sec:appendix_additional_experiments}). These results indicate that our CTSN and TMPR methods effectively advance the gradient-based SNN optimization. 

\begin{table}[!t]
  \centering
  \captionsetup{width=\columnwidth}
  \caption{Ablation study results of the proposed method (\%). w/ C: with CTSN. w/ T: with TMPR. w/ C\&T: with CTSN \& TMPR.}
  \label{tab:ablation_of_effectiveness}
  \begin{threeparttable}
    \begin{tabularx}{\columnwidth}{
    >{\hsize=1.7\hsize}C
    >{\hsize=0.3\hsize}C
    CCCC}
      \toprule
      &  & \multicolumn{4}{c}{\textbf{Method}}\\
      \cmidrule(l){3-6}
      \multirow{-2.5}{*}{\textbf{Dataset}}  & \multirow{-2.5}{*}{\textbf{T}} & Ternary & w/ C & w/ T & w/ C\&T \\
      \toprule
      CIFAR-100 & \makecell{4\\6} & \makecell{$73.95$\\ $74.12$} & \makecell{$74.18$\\ $74.48$} & \makecell{$74.57$\\ $74.73$} & \makecell{$74.82$\\ $75.12$} \\
      \midrule
      CIFAR10-DVS & 10 & $80.30$ & $80.57$ & $80.50$ & $81.23$\\
      \bottomrule
    \end{tabularx}
  \end{threeparttable}
\end{table}
\textbf{Influence of the Hyperparameter in TMPR Method.} To explore the influence of the regularization coefficient $\lambda$ on the performance of the proposed TMPR method, we conduct experiments on CIFAR-100 and CIFAR10-DVS datasets, taking values set to $\{0.01,0.05,0.1,0.25,0.5,1.0\}$. Results in Figure. \ref{fig:lambda} show that the performance of TMPR method first increases when $\lambda>0$, and then degrades significantly when $\lambda \ge 0.25$. We argue that this is due to the fact that too large $\lambda$ will cause excessive regularization in early timesteps, which dilutes the gradients generated by the classification loss $\mathcal{L}_{CE}$ during backpropagation, providing the network with an incorrect learning direction. As a result, we set $\lambda$ to $0.05$ for CIFAR-10/100 and ImageNet-100, $0.01$ for CIFAR10-DVS. The results provided in \ref{sec:appendix_additional_experiments} demonstrates that the optimal value of $\lambda$ for TMPR is not sensitive to the architecture, supporting the cross-architecture generalizability of TMPR.

\begin{figure}[!t]
  \centering
  \includegraphics[width=0.6\columnwidth]{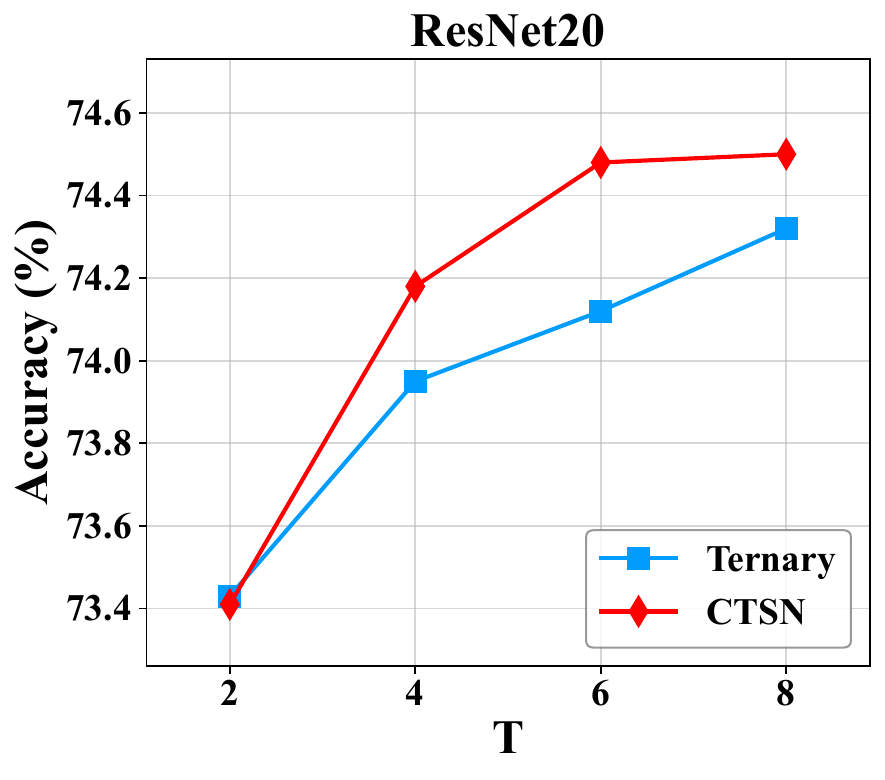}
  \captionsetup{width=\columnwidth}
  \caption{Timestep ablation study of CTSN.}
  \label{fig:temporal_ablation}
\end{figure}
\textbf{Temporal Ablation of CTSN.} To validate the effectiveness of CTSN on magnifying the temporal gradient, we verify the performance comparison between CTSN and the ternary spiking neuron with various timesteps. The experiment is conducted on CIFAR-100 with ResNet20 as the backbone. The results are demonstrated in Figure. \ref{fig:temporal_ablation}. When the total timestep is set to 2, the complemented term in our proposed CTSN model cannot magnify the temporal gradient and yield accuracy around $73.4\%$ similar to the ternary spiking neuron. As the timestep increases, CTSN consistently outperforms the ternary spiking neuron, indicating that the complemented term in CTSN contributes to the temporal gradient. This is consistent with the gradient analysis in Section. \ref{sec:dynamic_and_theoretical_analysis}.

\begin{figure}[!t]
  \centering
  \includegraphics[width=\columnwidth]{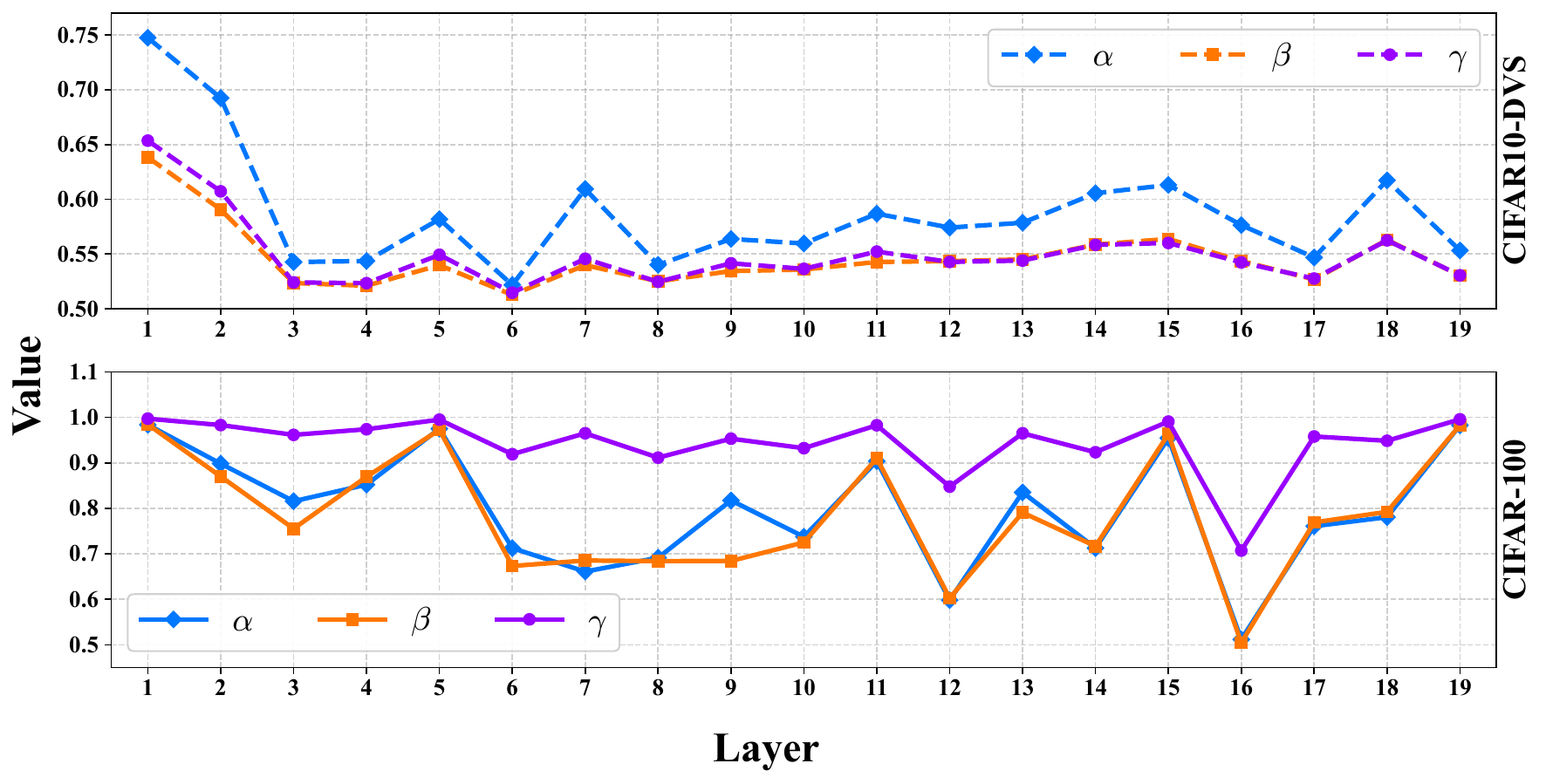}
  \captionsetup{width=\columnwidth}
  \caption{Tracking of learnable parameters in each layer of ResNet20. Upper: Network trained on CIFAR10-DVS with 10 timesteps. Lower: Network trained on CIFAR-100 with 4 timesteps.}
  \label{fig:learnable_parameters}
\end{figure}
\textbf{Learnable Parameters Tracking.} As shown in Eq. \ref{eq:g_static} and Eq. \ref{eq:g_neuromorphic}, our CTSN model contains three learnable parameters. We visulalize the learnable parameters of each layer in ResNet20 (without TMPR) on CIFAR-100 and CIFAR10-DVS datasets. As described in Section. \ref{sec:ctsn}, the initial values of $\alpha, \beta, \gamma$ are all set to $0.5$. As shown in Figure. \ref{fig:learnable_parameters}, $\alpha$, $\beta$ and $\gamma$ have undergone sufficient learning and adjustment on both datasets, confirming the effectiveness of parameter learnability and demonstrating the heterogeneity of neurons in the network.

\textbf{Membrane Potential Visualization.} We visualize the membrane potential distribution of CTSN neuron models in ResNet20 (trained by TMPR) on CIFAR-100 and CIFAR10-DVS datasets. As observed from Figure. \ref{fig:membrane_potential_distribution}, compared to ternary neurons, the distribution of CTSN varies more smoothly over time on both datasets, indicating more stable temporal gradients. Due to the use of the TMPR method for training, the membrane potential distribution demonstrates stronger compactness and symmetry, especially on CIFAR-100 dataset. These results demonstrate that the proposed CTSN and TMPR methods improve the dynamic characteristics of ternary neurons, effectively enhancing model performance by preventing over activation or excessive silence of neurons. Detailed analysis is provided in \ref{sec:appendix_membrane_visulization}.

\subsection{Comparison with SOTA Methods}
We evaluate CTSN using both static and neuromorphic datasets: CIFAR-10/100 \citep{krizhevsky2009cifar} and ImageNet-100 \citep{deng2009imagenet} are static datasets, while CIFAR10-DVS \citep{li2017cifar10dvs} is a neuromorphic dataset. For static datasets, we utilize direct encoding \citep{rathi2021diet} to transfer image data into time series. We apply normalization to all image data to make sure input images have 0 mean and 1 variance. We conduct random horizontal flipping and cropping, AutoAugment \citep{cubuk2019autoaugment} and Cutout \citep{devries2017cutout} on the training data of CIFAR-10/100. For ImageNet-100, all images were uniformly resized to 224×224 pixels prior to processing. We apply random horizontal flipping and cropping to training images, while the center cropping is applied to validation images. For neuromorphic dataset CIFAR10-DVS, we split the dataset at a 9:1 ratio into 9k training smaples and 1k test samples. All event frames are preprocessed and resized to 48×48 pixels. Following \citep{deng2022tet}, we take random horizontal flipping and random roll within 5 pixels as augmentation for training data. All the experiments are conducted on NVIDIA RTX 4090D GPU. We implement our methods based on PyTorch framework and utilize SGD optimizer with momentum set to 0.9. The initial learning rate is set to 0.1 and decayed by a cosine annealing schedule. All models are trained for 300 epochs with batch size set to 64.

\begin{table}[!t]
    \centering
    \captionsetup{width=\textwidth}
    \caption{Comparison with other state-of-the-art methods. We compare our method with the results reported in the respective literature. *: self-implementation results. $\dag$: implementation of our methods with open source code \citep{fang2023psn}.}
    \label{tab:comparison_static}
    \begin{threeparttable}
    \begin{tabularx}{\textwidth}{
    >{\hsize=0.5\hsize}C
    >{\hsize=1.7\hsize}C
    >{\hsize=1.3\hsize}C
    >{\hsize=0.3\hsize}C
    >{\hsize=1.2\hsize}C}
        \toprule
        \textbf{Dataset} & \textbf{Methods} & \textbf{Architecture} & \textbf{T} & \textbf{Accuracy (\%)}\\
        \toprule
            \multirow{11}{*}{\rotatebox[origin=c]{90}{\textbf{CIFAR-10}}}
            & PLIF \citeyearpar{fang2021plif} & PLIF Net & 6 & $94.25$\\
            & GLIF \citeyearpar{yao2022glif} & ResNet18 & 6 & $94.88$\\
            & TET \citeyearpar{deng2022tet} & ResNet19 & 4 & $94.44$\\
            & RMP-Loss \citeyearpar{guo2023rmp} & ResNet19 & 4 & $95.51$\\
            & ASGL \citeyearpar{wang2023asgl} & ResNet18 & 4 & $95.35$\\
            & CLIF \citeyearpar{huang2024clif} & ResNet18 & 4 & $96.01$\\
            & \multirow{1.8}{*}{Ternary Spike\tnote{*} \citeyearpar{guo2024ternary}} 
            & ResNet19 & 4 & $96.20$\\
            & & ResNet20 & 6 & $94.62$\\
            \cmidrule{2-5}
            & \multirow{3}{*}{\textbf{Ours}} 
            & \textbf{ResNet19} & \textbf{4} & $\mathbf{96.46}$\\
            \cmidrule{3-5}
            & & \textbf{ResNet20} & \makecell{\textbf{4} \\ \textbf{6}}& \makecell{$\mathbf{94.92}$ \\ $\mathbf{95.01}$}\\
        \midrule
            \multirow{10.5}{*}{\rotatebox[origin=c]{90}{\textbf{CIFAR-100}}}
            & GLIF \citeyearpar{yao2022glif} & ResNet18 & 6 & $77.35$\\
            & TET \citeyearpar{deng2022tet} & ResNet19 & 4 & $74.47$\\
            & RMP-Loss \citeyearpar{guo2023rmp} & ResNet19 & 4 & $78.28$\\
            & ASGL \citeyearpar{wang2023asgl} & ResNet18 & 4 & $77.74$\\
            & CLIF \citeyearpar{huang2024clif} & ResNet18 & 4 & $79.69$\\
            & \multirow{1.8}{*}{Ternary Spike\tnote{*}\citeyearpar{guo2024ternary}} 
            & ResNet19 & 4 & $79.68$\\
            & & ResNet20 & 6& $74.12$\\
            \cmidrule{2-5}
            & \multirow{3}{*}{\textbf{Ours}} 
            & \textbf{ResNet19} & \textbf{4} & $\mathbf{81.19}$\\
            \cmidrule{3-5}
            & & \textbf{ResNet20} & \makecell{\textbf{4} \\ \textbf{6}}& \makecell{$\mathbf{74.82}$ \\ $\mathbf{75.12}$}\\
        \midrule
            \multirow{5.5}{*}{\rotatebox[origin=c]{90}{\textbf{ImageNet100}}}
            & LocalZO \citeyearpar{mukhoty2023localzo} & SEW-ResNet34 & 4 & $81.56$\\
            & IMP+LTS \citeyearpar{shen2024implts} & SEW-ResNet34 & 4 & $80.80$\\
            & FPT \citeyearpar{feng2025fpt} & SEW-ResNet34 & 4 & $83.27$\\
            \cmidrule{2-5}
            & \multirow{2.4}{*}{\textbf{Ours}} 
            & \textbf{ResNet34} & \textbf{4} & $\mathbf{83.78}$\\
            \cmidrule{3-5}
            & & \textbf{SEW-ResNet34} & \textbf{4} &$\mathbf{85.06}$\\
        \bottomrule
    \end{tabularx}
    \end{threeparttable}
\end{table}
\begin{table}[!t]
    \centering
    \captionsetup{width=\textwidth}
    \caption{Comparison with the existing neuron models on CIFAR10-DVS dataset. * denotes self-implementation results. \dag\ denotes implementation of our methods directly in open source code provided by \citet{fang2023psn}.}
    \label{tab:comparison_neuromorphic}
    \begin{threeparttable}
    \begin{tabularx}{\textwidth}{
    >{\hsize=1.7\hsize}C
    C
    >{\hsize=0.3\hsize}C
    C}
        \toprule
        \textbf{Methods} & \textbf{Architecture} & \textbf{T} & \textbf{Accuracy (\%)}\\
        \toprule
            PLIF \citeyearpar{fang2021plif} & \makecell{PLIF Net\\ ResNet20\tnote{*}} & \makecell{20\\10} & \makecell{$74.80$\\$80.70$}\\
            GLIF \citeyearpar{yao2022glif} & \makecell{7B-wideNet\\ ResNet20\tnote{*}} & \makecell{16\\10} & \makecell{$78.10$\\$77.65$}\\
            CLIF \citeyearpar{huang2024clif} & \makecell{VGG11\\ ResNet20\tnote{*}} & \makecell{16\\10} & \makecell{$79.00$\\$79.53$}\\
            Ternary Spike \citeyearpar{guo2024ternary} & \makecell{ResNet20\\ResNet20\tnote{*}} & \makecell{10\\10} & \makecell{$79.80$\\$80.30$}\\
            \midrule
            \multirow{3.5}{*}{\textbf{Ours}} 
            & \textbf{VGG16} & \textbf{10} & $\mathbf{79.06}$\\
            \cmidrule{2-4}
            & \textbf{ResNet20} & \textbf{10}& $\mathbf{81.23}$\\
            \cmidrule{2-4}
            & \textbf{VGGSNN}\tnote{\dag} & \textbf{10}& $\mathbf{83.20}$\\
        \bottomrule
    \end{tabularx}
    \end{threeparttable}
\end{table}
Table. \ref{tab:comparison_static} reveals that our ternary SNN always outperforms the ternary spiking neuron and surpasses all other SNN methods within the same network backbone. Our approach achieves $96.46\%$ accuracy on CIFAR-10 and $81.19\%$ accuracy on CIFAR-100, outperforming other SNN models. On ImageNet-100 dataset, our approach achieves $85.06\%$ accuracy with 4 timesteps, which is a significant improvement compared to other methods. On the widely used neuromorphic dataset CIFAR10-DVS, we compare our approach with other state-of-the-art neuron models \citep{fang2021plif,yao2022glif,huang2024clif,guo2024ternary}, including both the results reported in the original studies and those obtained through self-implementation. As shown in Table. \ref{tab:comparison_neuromorphic}, our approach scores $79.06\%$ and $81.23\%$, which outperform other spiking neuron models. Notably, we implement our approach in open source code and achieve $83.20\%$ using VGGSNN, indicating its flexibility. 

It is worth noting that, to ensure consistency with our baseline method Ternary Spike \citep{guo2024ternary}, we adopt a lightweight ResNet20 architecture, while ResNet19 and ResNet34 follow the architectures commonly used in related works, both outperforming the baseline method.

We also conduct experiments to evaluate the computational overhead and energy consumption of our approach. Detailed statistics of computational cost, average spike firing rate and estimated energy consumption are described in \ref{sec:appendix_computational_overhead_and_energy_consumption}.

\section{Discussions}
\subsection{Bio-plausibility of CTSN}
\label{sec:biological}
The design of CTSN draws its inspiration primarily from the core characteristics of the biological nervous system across both spatial and temporal dimensions.

Spatially, cortical computation depends on a precise excitation–inhibition (E/I) ratio that maintains circuit homeostasis \citep{xue2014equalizing}. Its disruption impairs neural coding and behavior and may produce hyperexcitation or functional silence \citep{yizhar2011neocortical}. Interactions between excitatory pyramidal neurons and inhibitory GABAergic interneurons constrain synaptic integration through feedforward inhibition, preserving spike-timing precision and cortical spatial tuning \citep{isaacson2011inhibition}. At the representation level, binary SNNs encode only silence and excitation with $0$ and $1$, leaving inhibition implicit and potentially losing signed interlayer information. CTSN therefore adopts ternary spiking neurons \citep{guo2024ternary}, whose $1$, $0$, and $-1$ spikes explicitly represent excitation, silence, and inhibition, respectively. 

Temporally, cortical neurons exhibit hierarchical sustained activity without continuous input \citep{egorov2002graded} and heterogeneous timescales across regions and neurons \citep{wolff2022intrinsic}. Since biological responses depend on prior membrane states \citep{zhao2020membrane}, identical inputs may elicit different outputs, whereas hard resets erase this history after each spike. CTSN instead introduces a non-resetting adaptive memory term $h(t)$ that retains membrane potential across timesteps and adjusts memory strength, allowing representations to accumulate over time instead of restarting from zero.

\subsection{Application Prospects}
Both CTSN and TMPR impose no constraints on specific network architectures, ensuring their general applicability across various SNN models. The depolarization-like mechanism provided by the complemental term and the ternary spiking characteristic offers CTSN great biological plausibility, making CTSN a highly potential framework for modeling neural activities in computational neuroscience. The TMPR method is a practical and powerful approach for pretraining ternary SNNs, which can be utilized to provide high-performance pre-trained SNN models for various application scenarios.

\subsection{Limitations and Future Outlooks}
The underlying principles governing neuron dynamics and temporal information transmission mechanisms of both ternary spiking neuron and CTSN model are not yet fully understood. For different types of inputs, CTSN exhibits distinct neuron behaviors, making it a model lacks of scalability. Further studies on the underlying mechanisms and unified ternary spiking neuron models are required in the future. Due to the characteristic of emitting both positive and negative spikes, CTSN model generate spikes more frequently than binary neurons, resulting in higher energy consumption. Moreover, this characteristic leads to poor compatibility with existing neuromorphic hardware designed for binary neurons. Nevertheless, CTSN model and TMPR method introduce additional operations to membrane potentials, which inevitably reduce computational efficiency in training period. It is crucial to develop ternary spiking neuron models and training methods that offer higher computational and energy efficiency, as well as compatibility with neuromorphic hardware.

\section{Conclusion}
In this work, we investigate the problem of iterative information loss, temporal gradient vanishing, and the irregular membrane potential distribution of ternary spiking neurons. We further propose the CTSN model and TMPR method to address these issues. CTSN features additional complemental term in the integration process, which helps to magnify the temproal gradient during backpropagation, retains historical information, and smooths the membrane potential. TMPR method utilizes the complemented membrane potential as regularization, creating extra paths in gradient computation to enhance membrane potential aggregation, effectively narrowing the membrane potential distribution. Experimental results demonstrate that the CTSN model with TMPR method achieves state-of-the-art performance in various tasks with different network backbones. Furthermore, the CTSN model exhibits strong biological plausibility, offering great potentials in neuroscience. Nevertheless, due to the mathematical complexity of the CTSN model, a more comprehensive analysis of its neuron dynamic and temporal information transmission mechanisms remains necessary for future work.

\section*{Acknowledgment}
This work is supported in part by the National Natural Science Foundation of China under Grant 12401627, in part by the Open Research Fund of Guangdong Key Laboratory of Human Sports Performance Science under Grant GPKLHPS-OP-2026002, and in part by the Fundamental Research Funds for the Central Universities in China under Grant 2026JCYY02.

\appendix
\section{Performance of the Ternary Spiking Neuron with Soft Reset Mechanism}
\label{sec:appendix_soft_reset}
\begin{figure}[!t]
  \centering
  \includegraphics[width=0.8\textwidth]{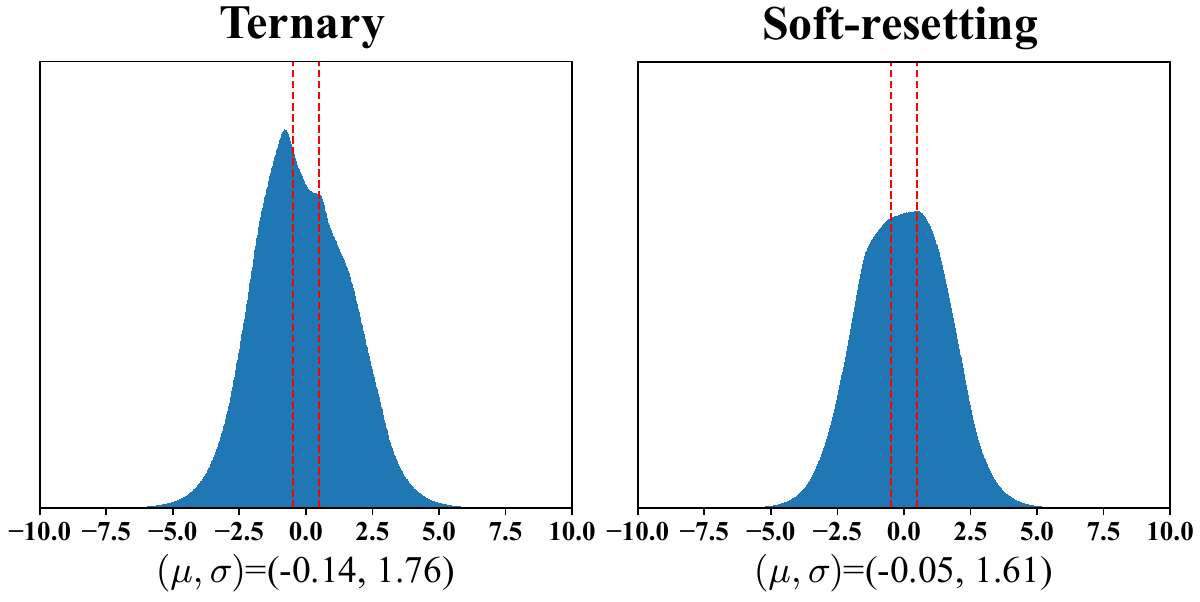}
  \captionsetup{width=\textwidth}
  \caption{Membrane potential Distributions for the third layer of ResNet20 across all timesteps. Left: Hard-resetting neuron. Right: Soft-resetting neuron. Dashed lines indicate the spiking thresholds.}
  \label{fig:appendix_mem_hard_soft_overall}
\end{figure}
As discussed in Section. \ref{sec:motivation_information_loss}, the soft reset mechanism is usually implemented to address the iterative information loss issue on binary spiking neurons. However, the experimental results in Table. \ref{tab:soft_reset} indicate that directly applying the soft reset mechanism to ternary spiking neurons is not effective. We argue that this is due to the fact that the ternary spiking neuron is capable of emitting both positive and negative spikes. For ResNet20 on CIFAR10-DVS, we visualize the membrane potential distribution of these two kind of ternary spiking neurons in Figure. \ref{fig:appendix_mem_hard_soft_overall} as an example. The spiking regions of its membrane potential ($(-\infty,-V_{th}]$ and $[V_{th},+\infty)$) are symmetric about zero. As shown in the left part of Figure. \ref{fig:appendix_mem_hard_soft_overall}, the distribution of the hard-resetting neurons is sharply peaked. Due to the hard reset mechanism, the membrane potential will always be reset to zero, which is the midpoint of its spiking regions. On the countrary, although the distribution of the soft-resetting neurons is smoother, the soft reset mechanism forces the neuron to retain the residual potentials, which destroys the symmetry of the membrane potential. For example, for a soft-resetting neuron with $V_{th}=0.5$, membrane potential $u(t)=2.5$ will be reset to $2.0$ at $t+1$, which will still trigger a spike, which may lead to excessive neural activation in subsequent stages, resulting in performance degradation.

As for CTSN model defined in Section. \ref{sec:ctsn}, we can derive the iterative equation of output spike sequence $o(t)$ as
\begin{equation}
  \begin{split}
    o(t)&=\Theta\big(\tilde{u}(t), V_{th}\big)\\
          &=\Theta\biggl(\mathbb{G}\big(h(t-1),\tau \tilde{u}(t-1)(1-|o(t-1)|),\alpha,\beta,\gamma\big)+x(t), V_{th}\biggr)\\
          &=\Theta\biggl(\mathbb{G}\big(h(t-1),\tau (h(t-1)+x(t-1))(1-|o(t-1)|),\alpha,\beta,\gamma\big)\\
          &\quad + x(t), V_{th}\biggr)\\
          &\quad \dots\\
  \end{split}
  \label{eq:appendix_iterative_analysis}
\end{equation}
The iterative derivation of output spike sequence $o(t)$ essentially corresponds to the repeated invocation of function $\mathbb{G}\big(h(t-1),u(t),\alpha,\beta,\gamma\big)$. 
As can be seen from Eq. \ref{eq:appendix_iterative_analysis}, the iterative information loss caused by the hard reset mechanism $\hat{s}(t)$ persists. However, the newly introduced additional term $h(t)$ can be seen as a memristor without any reset mechanism. For those neurons that intermittently emit spikes, $h(t)$ can effectively retain information from inputs arrived in previous timesteps, thereby partially addressing the iterative information loss problem while preserving the symmetry of the reset mechanism.

\section{BPTT for Ternary Spiking Neuron with Surrogate Gradient}
\label{sec:appendix_gradient_ternary}
This section is mainly referenced from \citep{wu2018bptt,meng2023sltt,huang2024clif}. Firstly, we recall the ternary spiking neuron model defined in Eq. \ref{eq:ternary_u} - \ref{eq:ternary_fire} and rewrite this model as
\begin{equation}
  u^l(t)=\tau u^l(t-1)(1-|o^l(t-1)|)+\sum_{l=1}^{L}\mathbf{W}^lo^{l-1}(t)+b^l,
  \label{eq:appendix_ternary_u}
\end{equation}
and
\begin{equation}
  \begin{aligned}
    o^l(t) & = \Theta(u^l(t),V_{th})\\
    & =
    \begin{cases}
      1,  & \text{if }u^l(t)\ge V_{th} \\
      -1, & \text{if }u^l(t)\le -V_{th} \\
      0,  & otherwise                \\
    \end{cases} \\
  \end{aligned}.
  \label{eq:appendix_ternary_fire}
\end{equation}
As described in Section. \ref{sec:motivation_gradient}, the gradient of loss function $\mathcal{L}$ with respect to weight matrix $\mathbf{W}^l$ is calculated as
\begin{equation}
  \frac{\partial \mathcal{L}}{\partial \mathbf{W}^l(t)} = \sum_{t=1}^{T}\frac{\partial \mathcal{L}}{\partial u^l(t)}\frac{\partial u^l(t)}{\partial \mathbf{W}^l(t)},
  \label{eq:appendix_ternary_gradient}
\end{equation}
where $\mathcal{L}$ represents the loss function. The partial derivative $\frac{\partial \mathcal{L}}{\partial u^l(t)}$ can be recursively calculated as
\begin{equation}
  \frac{\partial \mathcal{L}}{\partial u^l(t)} = \frac{\partial \mathcal{L}}{\partial o^l(t)}\frac{\partial o^l(t)}{\partial u^l(t)} + \frac{\partial \mathcal{L}}{\partial u^l(t+1)}\epsilon^l(t),
  \label{eq:appendix_ternary_partial_derivative}
\end{equation}
where $\epsilon^l(t)$ is defined as
\begin{equation}
  \epsilon^l(t) \equiv \frac{\partial u^l(t+1)}{\partial u^l(t)} + \frac{\partial u^l(t+1)}{\partial |o^l(t)|}\frac{\partial |o^l(t)|}{\partial u^l(t)}.
\end{equation}
In this case, we can recursively calculate the gradient term $\frac{\partial \mathcal{L}}{\partial u^l(t)}$ as Eq. \ref{eq:appendix_gradient}, which is consistent with Eq. \ref{eq:gradient}.
\begin{equation}
  \begin{split}
    \frac{\partial \mathcal{L}}{\partial u^l(t)} &= \frac{\partial \mathcal{L}}{\partial o^l(t)}\frac{\partial o^l(t)}{\partial u^l(t)} + \frac{\partial \mathcal{L}}{\partial u^l(t+1)}\epsilon^l(t)\\
    &= \frac{\partial \mathcal{L}}{\partial o^l(t)}\frac{\partial o^l(t)}{\partial u^l(t)} + \biggl(\frac{\partial \mathcal{L}}{\partial o^l(t+1)}\frac{\partial o^l(t+1)}{\partial u^l(t+1)} + \frac{\partial \mathcal{L}}{\partial u^l(t+2)}\epsilon^l(t+1)\biggr)\epsilon^l(t)\\
    &= \frac{\partial \mathcal{L}}{\partial o^l(t)}\frac{\partial o^l(t)}{\partial u^l(t)} + \frac{\partial \mathcal{L}}{\partial o^l(t+1)}\frac{\partial o^l(t+1)}{\partial u^l(t+1)}\epsilon^l(t) + \frac{\partial \mathcal{L}}{\partial u^l(t+2)}\epsilon^l(t+1)\epsilon^l(t)\\
    &= \frac{\partial \mathcal{L}}{\partial o^l(t)}\frac{\partial o^l(t)}{\partial u^l(t)} + \frac{\partial \mathcal{L}}{\partial o^l(t+1)}\frac{\partial o^l(t+1)}{\partial u^l(t+1)}\epsilon^l(t) + \biggl(\frac{\partial \mathcal{L}}{\partial o^l(t+2)}\frac{\partial o^l(t+2)}{\partial u^l(t+2)}\\
    &\quad + \frac{\partial \mathcal{L}}{\partial u^l(t+3)}\epsilon^l(t+2)\biggr)\epsilon^l(t+1)\epsilon^l(t)\\
    &= \frac{\partial \mathcal{L}}{\partial o^l(t)}\frac{\partial o^l(t)}{\partial u^l(t)} + \frac{\partial \mathcal{L}}{\partial o^l(t+1)}\frac{\partial o^l(t+1)}{\partial u^l(t+1)}\epsilon^l(t)\\
    &\quad + \frac{\partial \mathcal{L}}{\partial o^l(t+2)}\frac{\partial o^l(t+2)}{\partial u^l(t+2)}\epsilon^l(t+1)\epsilon^l(t) + \frac{\partial \mathcal{L}}{\partial u^l(t+3)}\epsilon^l(t+2)\epsilon^l(t+1)\epsilon^l(t)\\
    &\quad \dots \\
    &= \frac{\partial \mathcal{L}}{\partial o^l(t)}\frac{\partial o^l(t)}{\partial u^l(t)}+\sum_{t'=t+1}^{T}\frac{\partial \mathcal{L}}{\partial o^l(t')}\frac{\partial o^l(t')}{\partial u^l(t')}\prod_{t''=1}^{t'-t}\epsilon^l(t'-t'')
    \end{split}
  \label{eq:appendix_gradient}
\end{equation}
When $t=T$, the recursive part does not exist and Eq. \ref{eq:appendix_gradient} reduces to
\begin{equation}
  \frac{\partial \mathcal{L}}{\partial u^l(T)} = \frac{\partial \mathcal{L}}{\partial o^l(T)}\frac{\partial o^l(T)}{\partial u^l(T)}.
\end{equation}
For intermediate layers $l=L-1,L-2,\dots,1$ in SNN, the gradient term $\frac{\partial \mathcal{L}}{\partial u^l(t)}$ is obtained from the previous layer during backpropagation. Therefore, $\frac{\partial \mathcal{L}}{\partial u^l(t)}$ can be rewritten as
\begin{equation}
  \begin{split}
    \frac{\partial \mathcal{L}}{\partial u^l(t)}&=\sum_{t'=t+1}^{T}\frac{\partial \mathcal{L}}{\partial u^{l+1}(t')}\frac{\partial u^{l+1}(t')}{\partial o^l(t')}\frac{\partial o^l(t')}{u^l(t')}\prod_{t''=1}^{t'-t}\epsilon^l(t'-t'')\\
    &\quad +\frac{\partial \mathcal{L}}{\partial u^{l+1}(t)}\frac{\partial u^{l+1}(t)}{\partial o^l(t)}\frac{\partial o^l(t)}{u^l(t)}.
  \end{split}
  \label{eq:appendix_gradient_intermediate}
\end{equation}
As for the last layer $l=L$ of SNN, the gradient term $\frac{\partial \mathcal{L}}{\partial u^L(t)}$ is obtained directly from the loss function during backpropagation. Therefore, $\frac{\partial \mathcal{L}}{\partial u^L(t)}$ remains consistent with Eq. \ref{eq:appendix_gradient}:
\begin{equation}
  \begin{split}
      \frac{\partial \mathcal{L}}{\partial u^L(t)}&=\frac{\partial \mathcal{L}}{\partial o^L(t)}\frac{\partial o^L(t)}{\partial u^L(t)}\\
      &\quad +\sum_{t'=t+1}^{T}\frac{\partial \mathcal{L}}{\partial o^L(t')}\frac{\partial o^L(t')}{\partial u^L(t')}\prod_{t''=1}^{t'-t}\epsilon^L(t'-t'').
  \end{split}
  \label{eq:appendix_gradient_last}
\end{equation}
The Eq. \ref{eq:appendix_gradient_intermediate} and Eq. \ref{eq:appendix_gradient_last} corresponds to Eq. \ref{eq:gradient} and Eq. \ref{eq:gradient_layers}.

\section{Visualization of the Membrane Potential}
\label{sec:appendix_membrane_visulization}
\begin{figure}[!t]
  \centering
  \subfloat[]{\includegraphics[width=0.8\columnwidth]{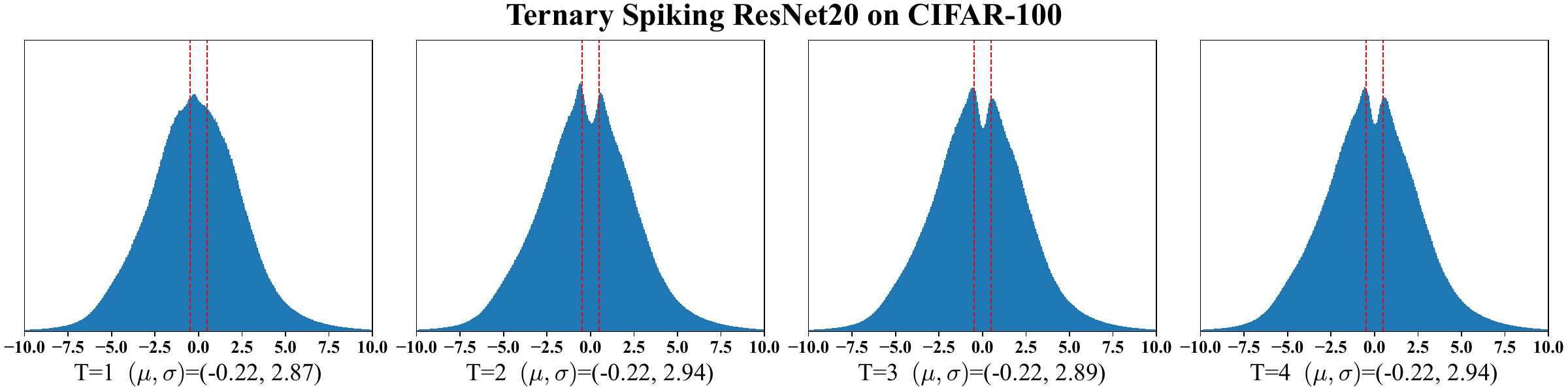}}\\
  \subfloat[]{\includegraphics[width=0.8\columnwidth]{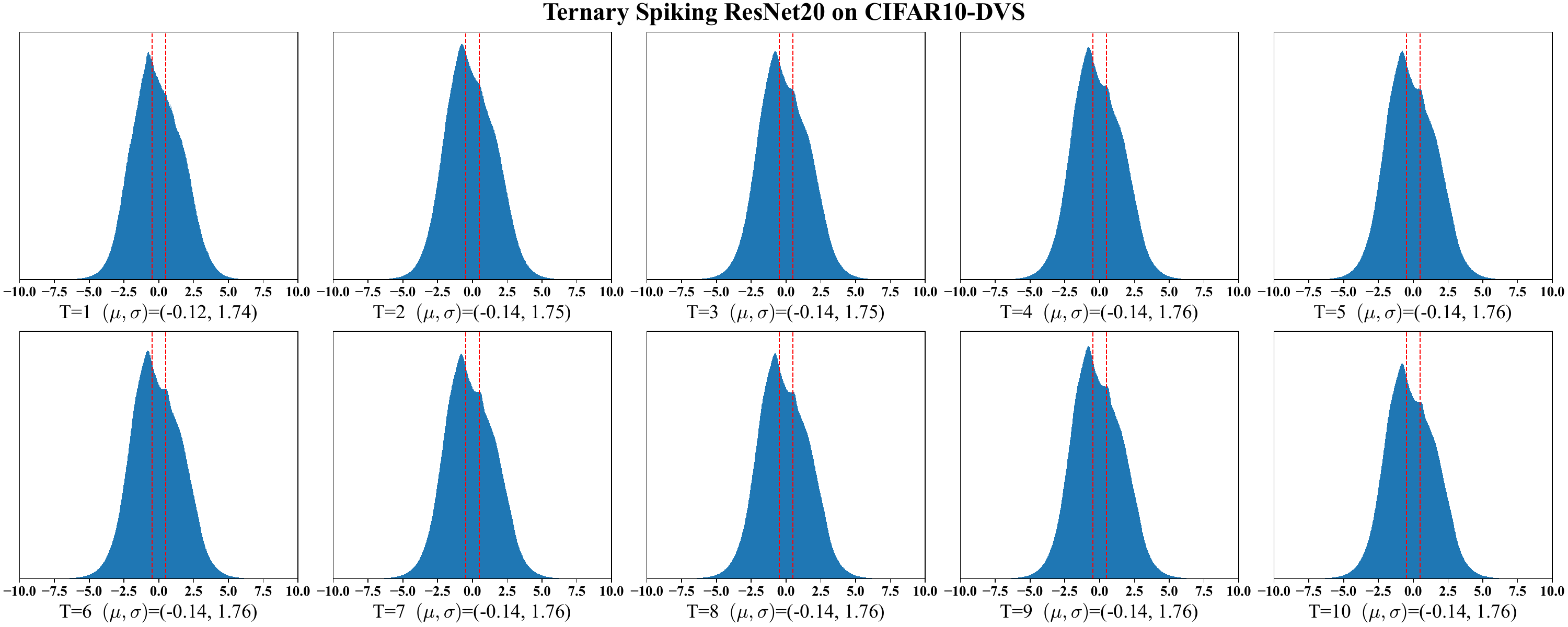}}
  \caption{Membrane potential Distributions for the third layer of Ternary Spiking ResNet20. (a) CIFAR-100. (b) CIFAR10-DVS.}
  \label{fig:appendix_mem_teranry}
\end{figure}
\begin{figure}[!t]
  \centering
  \subfloat[]{\includegraphics[width=0.8\columnwidth]{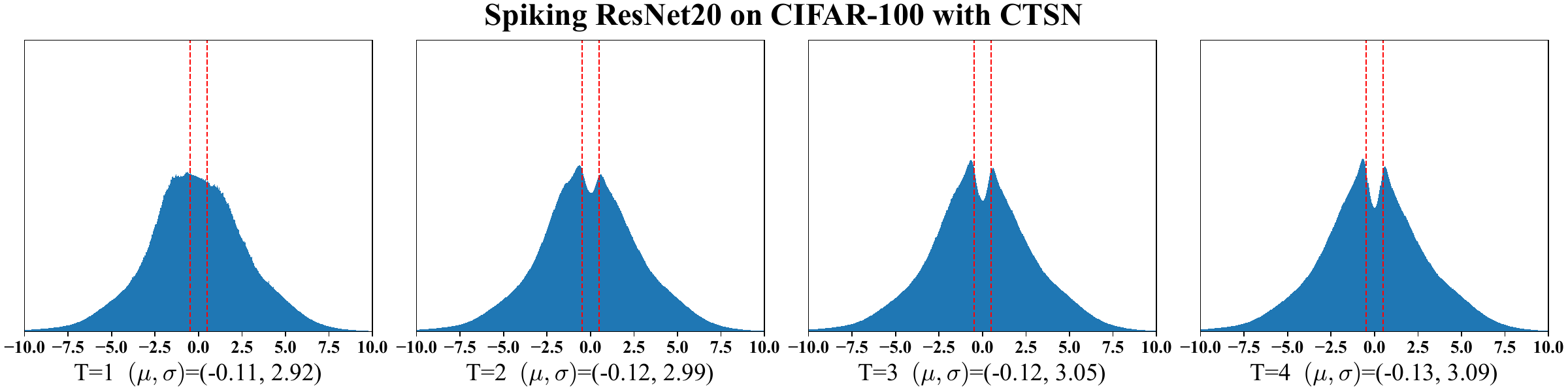}}\\
  \subfloat[]{\includegraphics[width=0.8\columnwidth]{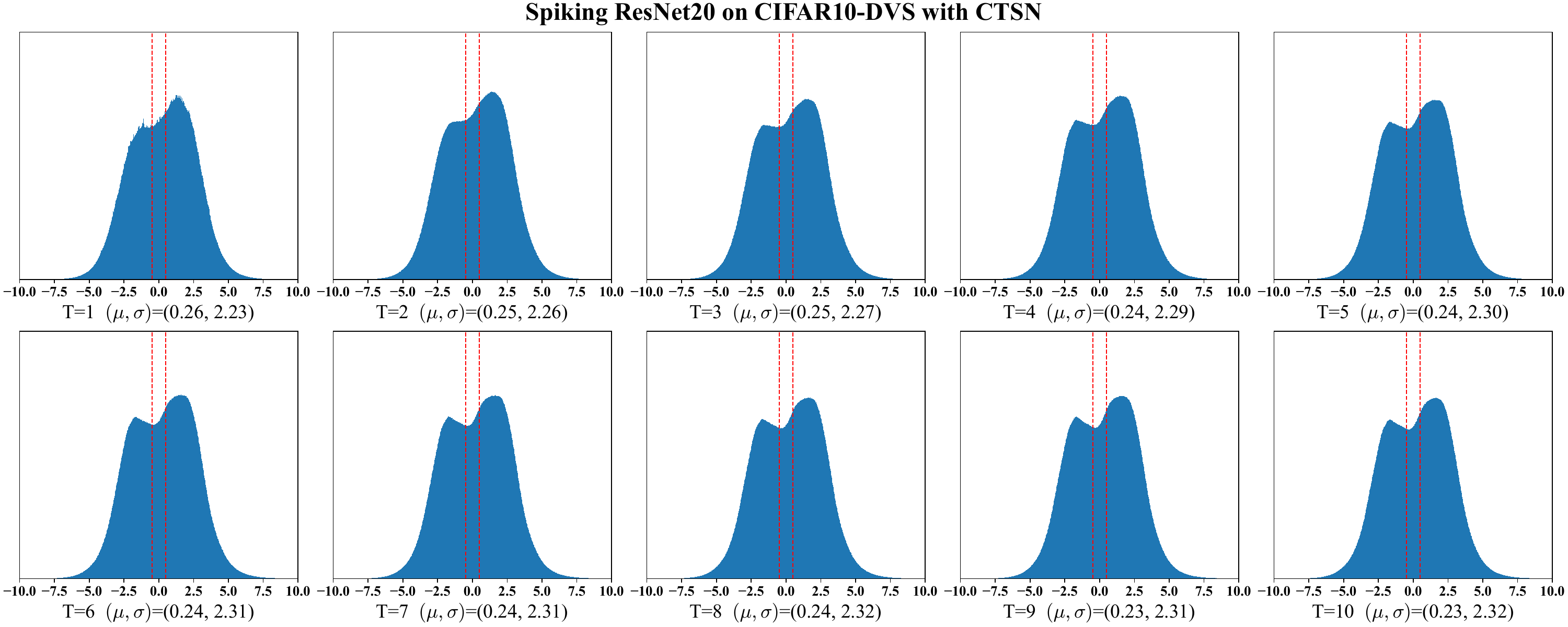}}
  \caption{Membrane potential Distributions for the third layer of CTSN Spiking ResNet20. (a) CIFAR-100. (b) CIFAR10-DVS.}
  \label{fig:appendix_mem_ctsn}
\end{figure}
\begin{figure}[!t]
  \centering
  \subfloat[]{\includegraphics[width=0.8\columnwidth]{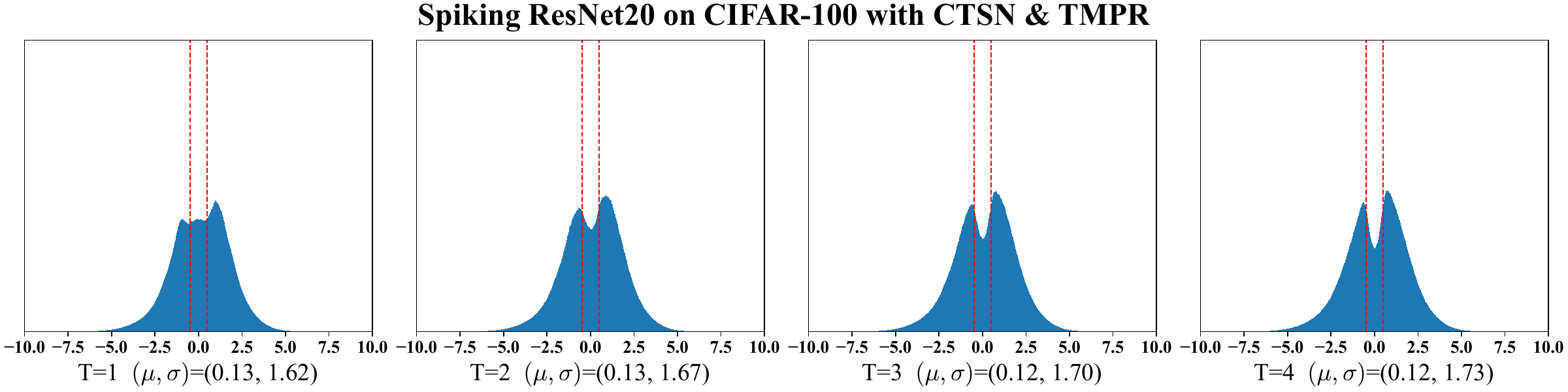}}\\
  \subfloat[]{\includegraphics[width=0.8\columnwidth]{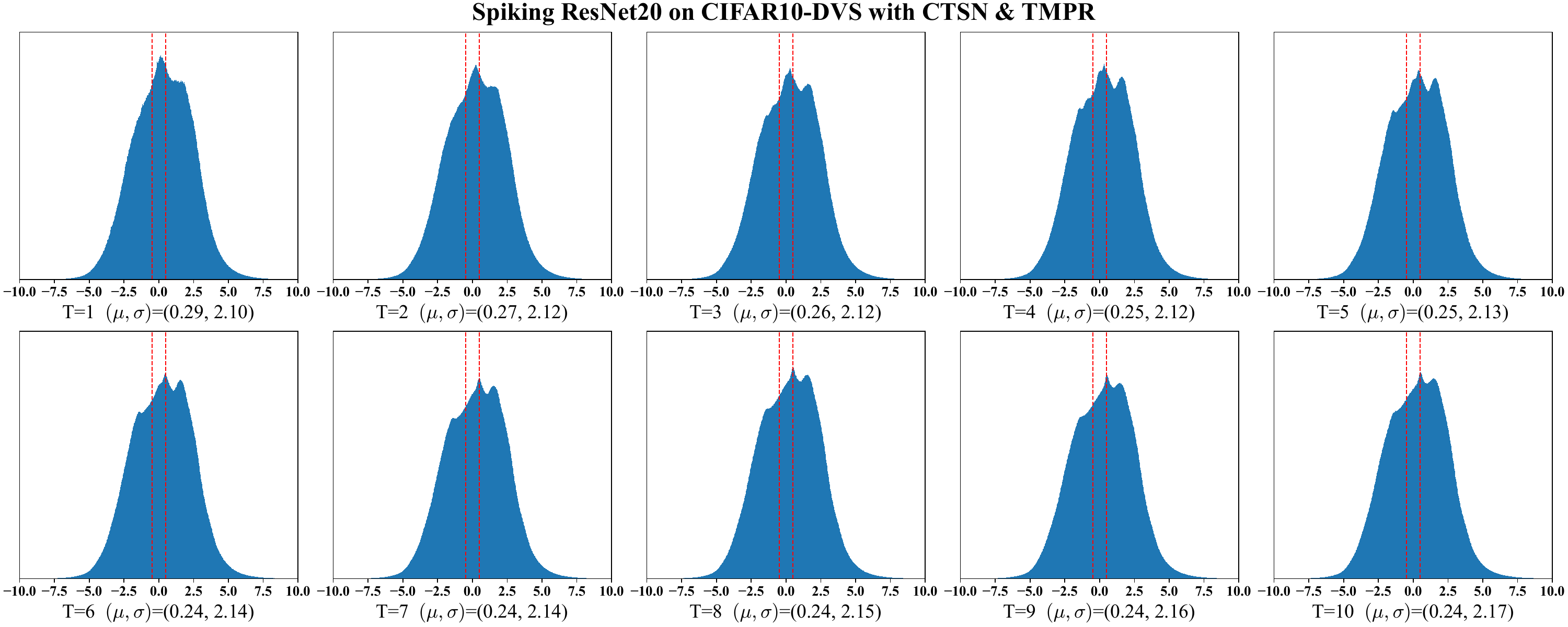}}
  \caption{Membrane potential Distributions for the third layer of CTSN Spiking ResNet20. (a) CIFAR-100 trained by TMPR. (b) CIFAR10-DVS.}
  \label{fig:appendix_mem_ctsn_tmpr}
\end{figure}
In this section, we provide detailed visualization of the membrane potential distributions for ResNet20. As observed from the Figure. \ref{fig:appendix_mem_teranry}, the distribution of ternary spiking neurons gradually derives into a bimodal pattern when processing static image data, but maintains sharply peaked unimodal distribution on neuromorphic data. As shown in Figure. \ref{fig:appendix_mem_ctsn}, the distribution of CTSN models varies more smoothly over time. On both CIFAR-100 dataset and CIFAR10-DVS dataset, the distributions demonstrate the trend of deriving into bimodal patterns on the first timestep, which is earlier than ternary spiking neurons, indicating more stable temporal gradients. However, the standard deviation of distribution on each timestep is larger than ternary spiking neurons, indicating server over activation. With the implementation of TMPR (as shown in Figure. \ref{fig:appendix_mem_ctsn_tmpr}), the distribution becomes narrower and increasingly concentrated toward gradient-active regions described in Eq. \ref{eq:part-II}. Furthermore, the membrane potential distribution exhibits stronger compactness and symmetry, especially on CIFAR-100 dataset. These results indicate that the proposed CTSN and TMPR methods effectively improve neuron dynamics, preventing membrane potentials from excessively residing in activation regions far from the threshold or in silent regions near zero, thereby avoiding over activation or excessive silence.

\section{The Gradients of CTSN}
\label{sec:appendix_gradient_ctsn}
The proposed CTSN model defined by Eq. \ref{eq:ctsn_u} - \ref{eq:g_neuromorphic} can be rewritten as
\begin{equation}
  h^l(t)=\mathbb{G}\biggl(h^l(t-1),\tau \tilde{u}^l(t-1)(1-|o^l(t-1)|),\alpha^l,\beta^l,\gamma^l\biggr),
  \label{eq:appendix_ctsn_u}
\end{equation}
\begin{equation}
  \tilde{u}^l(t)=h^l(t)+\sum_{l=1}^{L}\mathbf{W}^lo^{l-1}(t)+b^l,
  \label{eq:appendix_ctsn_tilde_u}
\end{equation}
and
\begin{equation}
  o^l(t) = \Theta(\tilde{u}^l(t),V_{th}).
\end{equation}
\begin{equation}
  \begin{split}
    \frac{\partial \mathcal{L}}{\partial \tilde{u}^l(t)} &= \frac{\partial \mathcal{L}}{\partial o^l(t)}\frac{\partial o^l(t)}{\partial \tilde{u}^l(t)} + \frac{\partial \mathcal{L}}{\partial \tilde{u}^l(t+1)}\underbrace{\frac{\partial \tilde{u}^l(t+1)}{\partial h^l(t+1)}}_{=1}\xi^l(t)\\
    &= \frac{\partial \mathcal{L}}{\partial o^l(t)}\frac{\partial o^l(t)}{\partial \tilde{u}^l(t)} + \frac{\partial \mathcal{L}}{\partial \tilde{u}^l(t+1)}\xi^l(t)\\
    &= \frac{\partial \mathcal{L}}{\partial o^l(t)}\frac{\partial o^l(t)}{\partial \tilde{u}^l(t)} + \biggl(\frac{\partial \mathcal{L}}{\partial o^l(t+1)}\frac{\partial o^l(t+1)}{\partial \tilde{u}^l(t+1)}\\
    &\quad + \frac{\partial \mathcal{L}}{\partial \tilde{u}^l(t+2)}\underbrace{\frac{\partial \tilde{u}^l(t+2)}{\partial h^l(t+2)}}_{=1}\big(\xi^l(t+1)+\frac{\partial h^l(t+2)}{\partial h^l(t+1)}\big)\biggr)\xi^l(t)\\
    &= \frac{\partial \mathcal{L}}{\partial o^l(t)}\frac{\partial o^l(t)}{\partial \tilde{u}^l(t)} + \frac{\partial \mathcal{L}}{\partial o^l(t+1)}\frac{\partial o^l(t+1)}{\partial \tilde{u}^l(t+1)}\xi^l(t)\\
    &\quad + \frac{\partial \mathcal{L}}{\partial \tilde{u}^l(t+2)}\big(\xi^l(t+1)+\nabla_{h^l}\mathbb{G}^l\big)\xi^l(t)\\
    &= \frac{\partial \mathcal{L}}{\partial o^l(t)}\frac{\partial o^l(t)}{\partial \tilde{u}^l(t)} + \frac{\partial \mathcal{L}}{\partial o^l(t+1)}\frac{\partial o^l(t+1)}{\partial \tilde{u}^l(t+1)}\xi^l(t) + \biggl(\frac{\partial \mathcal{L}}{\partial o^l(t+2)}\frac{\partial o^l(t+2)}{\partial \tilde{u}^l(t+2)}\\
    &\quad + \frac{\partial \mathcal{L}}{\partial \tilde{u}^l(t+3)}\underbrace{\frac{\tilde{u}^l(t+3)}{h^l(t+3)}}_{=1}\big(\xi^l(t+2)\\ 
    &\quad + \frac{\partial h(t+3)}{\partial h(t+2)}\big)\biggr)\cdot\big(\xi^l(t+1)+\nabla_{h^l}\mathbb{G}^l\big)\xi^l(t)\\
    &= \frac{\partial \mathcal{L}}{\partial o^l(t)}\frac{\partial o^l(t)}{\partial \tilde{u}^l(t)} + \frac{\partial \mathcal{L}}{\partial o^l(t+1)}\frac{\partial o^l(t+1)}{\partial \tilde{u}^l(t+1)}\xi^l(t)\\
    &\quad + \frac{\partial \mathcal{L}}{\partial o^l(t+2)}\frac{\partial o^l(t+2)}{\partial \tilde{u}^l(t+2)}\big(\xi^l(t+1)+\nabla_{h^l}\mathbb{G}^l\big)\xi^l(t)\\
    &\quad + \frac{\partial \mathcal{L}}{\partial \tilde{u}^l(t+3)}\frac{\tilde{u}^l(t+3)}{h^l(t+3)}\big(\xi^l(t+2)+\nabla_{h^l}\mathbb{G}^l\big)\big(\xi^l(t+1)+\nabla_{h^l}\mathbb{G}^l\big)\xi^l(t)\\
    &\quad \dots \\
    &= \frac{\partial \mathcal{L}}{\partial o^l(t)}\frac{\partial o^l(t)}{\partial \tilde{u}^l(t)}+\frac{\partial \mathcal{L}}{\partial o^l(t+1)}\frac{\partial o^l(t+1)}{\partial \tilde{u}^l(t+1)}\xi^l(t)\\
    &\quad + \sum_{t'=t+2}^{T}\big(\frac{\partial \mathcal{L}}{\partial o^l(t')}\frac{\partial o^l(t')}{\partial \tilde{u}^l(t')}\prod_{t''=1}^{t'-t}(\xi^l(t'-t'')+\underbrace{\nabla_{h^l}\mathbb{G}^l}_{\substack{\text{Complemental} \\ \text{Gradient}}})\big)
  \end{split}
  \label{eq:appendix_gradient_ctsn}
\end{equation}
The function $\mathbb{G}$ for static image data can be rewritten as
\begin{equation}
  \begin{split}
    \mathbb{G}(h(t-1),u,\alpha,\beta,\gamma)&=\alpha \text{ReLU}(h(t-1))+\beta(-\text{ReLU}(-h(t-1)))+\gamma u.
  \end{split}
  \label{eq:appendix_g_static}
\end{equation}
The function $\mathbb{G}$ for neuromorphic data can be rewritten as
\begin{equation}
  \begin{split}
    \mathbb{G}(h(t-1),u,\alpha,\beta,\gamma)&=\alpha h(t-1)+\beta \text{ReLU}(u)+\gamma (-\text{ReLU}(-u)).
  \end{split}
  \label{eq:appendix_g_neuromorphic}
\end{equation}
We define function $\text{Choice}(x,v_{th},a,b)$ as
\begin{equation}
  \begin{aligned}
    \text{Choice}(x,v_{th},a,b)
    & = 
    \begin{cases}
      a, & x \geq v_{th} \\
      b, & x < v_{th} \\
    \end{cases}
  \end{aligned}.
\end{equation}
The gradient term $\frac{\partial \mathcal{L}}{\partial \tilde{u}^l(t)}$ can be recursively calculated as
\begin{equation}
  \frac{\partial \mathcal{L}}{\partial \tilde{u}^l(t)} = \frac{\partial \mathcal{L}}{\partial o^l(t)}\frac{\partial o^l(t)}{\partial \tilde{u}^l(t)} + \frac{\partial \mathcal{L}}{\partial \tilde{u}^l(t+1)}\frac{\partial \tilde{u}^l(t+1)}{\partial h^l(t+1)}\xi^l(t),
  \label{eq:appendix_gradient_l_u}
\end{equation}
where $\xi^l(t)$ is expanded as
\begin{equation}
  \begin{split}
    \xi^l(t) &= \frac{\partial h^l(t+1)}{\partial \tilde{u}^l(t)}+\frac{\partial h^l(t+1)}{\partial |o(t)|}\frac{\partial |o(t)|}{\partial \tilde{u}^l(t)}\\
    &= \nabla_{\tilde{u}^l}\mathbb{G}^l+\frac{\partial h^l(t+1)}{\partial |o(t)|}\frac{\partial |o(t)|}{\partial \tilde{u}^l(t)}.
  \end{split}
  \label{eq:appendix_xi}
\end{equation}
From Eq. \ref{eq:appendix_g_static} and Eq. \ref{eq:appendix_g_neuromorphic}, $\nabla_{\tilde{u}^l}\mathbb{G}^l$ is calculated as
\begin{equation}
  \begin{aligned}
    \nabla_{\tilde{u}^l}\mathbb{G}^l
    & =
    \begin{cases}
      \gamma^l,  & \text{static} \\
      \text{Choice}(\tilde{u}^l,0,\beta^l,\gamma^l), & \text{neuromorphic} \\
    \end{cases} \\
  \end{aligned}.
\end{equation}
\begin{figure}[!t]
  \centering
  \includegraphics[width=\textwidth]{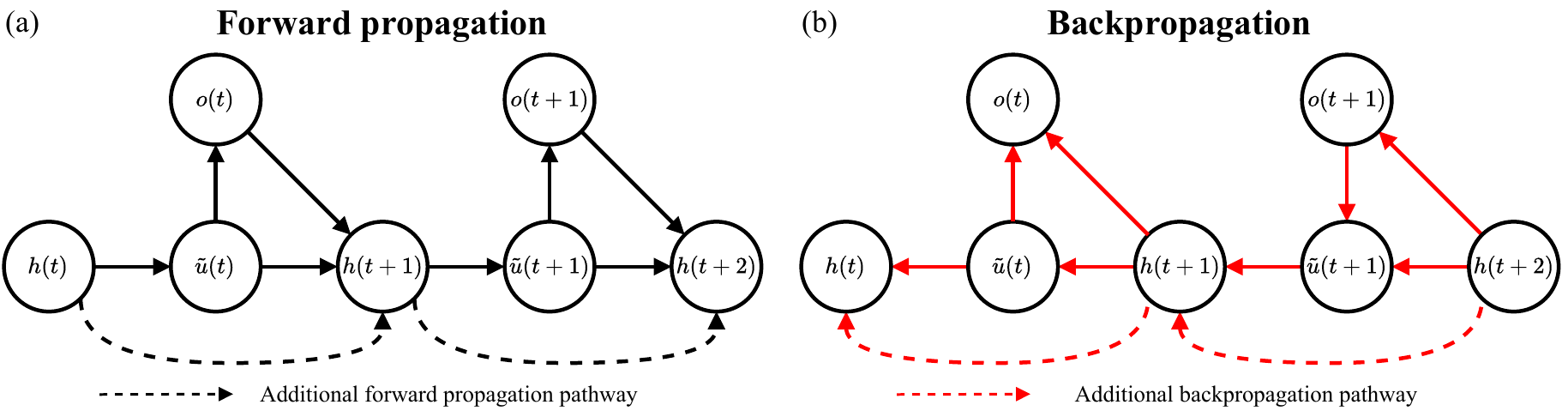}
  \captionsetup{width=\textwidth}
  \caption{The computation graph of (a) forward propagation and (b) backpropagation for CTSN model.}
  \label{fig:appendix_computation_graph}
\end{figure}
To better understand Eq. \ref{eq:appendix_gradient_l_u}, Eq. \ref{eq:appendix_xi} and the following derivation, we refer to Figure. \ref{fig:appendix_computation_graph}. In this case, we can recursively calculate the gradient term $\frac{\partial \mathcal{L}}{\partial \tilde{u}^l(t)}$ as Eq. \ref{eq:appendix_gradient_ctsn}, where $\nabla_{h^l}\mathbb{G}^l$ can be calculated as
\begin{equation}
  \begin{aligned}
    \nabla_{h^l}\mathbb{G}^l
    & =
    \begin{cases}
      \text{Choice}(h^l,0,\alpha^l,\beta^l),  & \text{static} \\
      \alpha^l, & \text{neuromorphic} \\
    \end{cases} \\
  \end{aligned}.
\end{equation}
Similar to the ternary spiking neuron, in CTSN model we have
\begin{equation}
  \frac{\partial \mathcal{L}}{\partial o^l(t)}=
  \begin{cases}
    \frac{\partial \mathcal{L}}{\partial o^{L}(t)}, & l=L \\
    \frac{\partial \mathcal{L}}{\partial u^{l+1}(t)}\frac{\partial u^{l+1}(t)}{\partial o^l(t)}, & L=L-1,L-2 \dots 1 \\
  \end{cases}.
  \label{eq:appendix_gradient_layers_ctsn}
\end{equation}
From Eq. \ref{eq:appendix_gradient_ctsn}, we can intuitively see that the additional term of CTSN magnifies the temporal gradient, which akin to taking a shortcut on the temporal dimension, mitigating the temporal gradient vanishing issue caused by the product operations.

\section{Further Validation of CTSN and TMPR on Complex Network}
\label{sec:appendix_additional_experiments}
\begin{figure}[!t]
  \centering
  \includegraphics[width=0.6\textwidth]{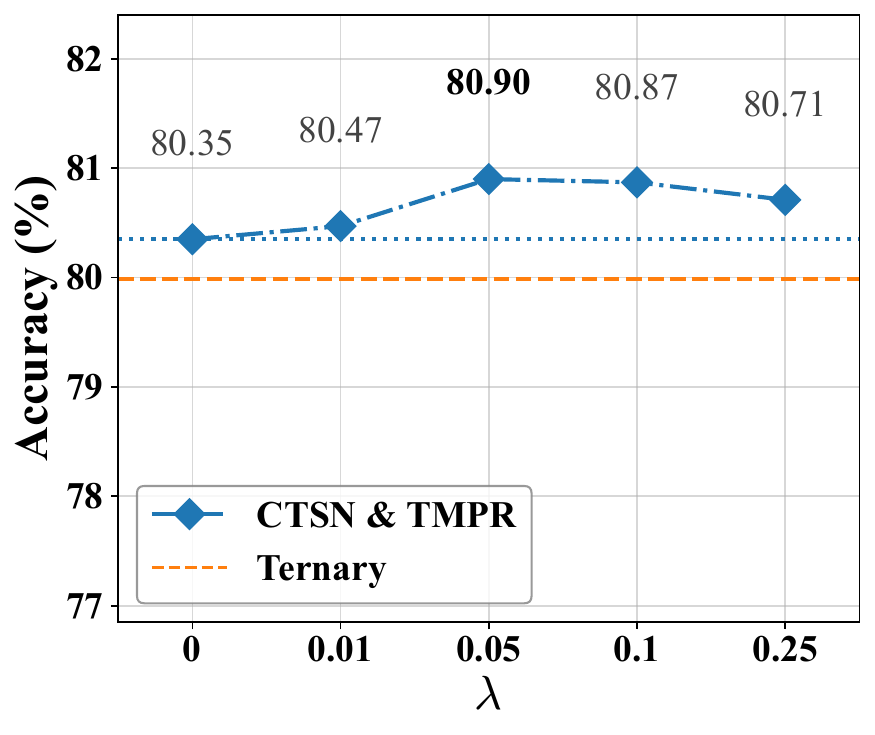}
  \captionsetup{width=\textwidth}
  \caption{Influence of the hyperparameter $\lambda$ on ResNet32. The orange dash line indicates the performance of the baseline ternary spiking network.}
  \label{fig:appendix_lambda}
\end{figure}
To further validate the effectiveness of our proposed methods and assess TMPR's sensitivity to the hyperparameter $\lambda$ in deeper networks, we conduct experiments on CIFAR100 utilizing commonly used ResNet32 as backbone, taking $\lambda$ values set to $\{0.01, 0.05, 0.1, 0.25\}$ and $0$ (disabling TMPR, CTSN only). The results are visualized in Figure. \ref{fig:appendix_lambda}. As can be observed, our proposed approaches achieved performance comparable to the state-of-the-art methods presented in Section. \ref{sec:experimental_results}, supporting the effectiveness of CTSN and TMPR. For ResNet32, the optimal value of $\lambda$ was 0.05, identical to that for ResNet20.

\section{Evaluation of Computational and Energy Costs}
\label{sec:appendix_computational_overhead_and_energy_consumption}
\subsection{Comparison of Computational Overhead}
\label{sec:appendix_computational_overhead}
\begin{table}[!t]
    \centering
    \caption{Comparison of training time, memory usage, and inference time. w/ C: with CTSN. w/ C\&T: with CTSN \& TMPR.}
    \label{tab:appendix_computational_overhead}
    \begin{threeparttable}
        \begin{tabularx}{\textwidth}{
            >{\hsize=0.9\hsize}C
            >{\hsize=0.6\hsize}C
            >{\hsize=1.1\hsize}C
            >{\hsize=1.2\hsize}C
            >{\hsize=1.2\hsize}C
        }
            \toprule
            \textbf{Dataset} & \textbf{Method} & \textbf{Training Time (hours)}
                    & \textbf{Memory Usage (MB)} & \textbf{Inference Time (s)} \\
            \midrule
            \multirow{3}{*}{\makecell{CIFAR100\\(T=4)}}
                & Ternary & 5.50 & 2213 & 3.55 \\
                & w/ C & 6.48 & 3087 & 3.84 \\
                & w/ C\&T & 8.75 & 4041 & 3.84 \\
            \midrule
            \multirow{3}{*}{\makecell{CIFAR10-DVS\\(T=10)}}
                & Ternary & 3.56 & 11381 & 2.02 \\
                & w/ C & 4.63 & 17975 & 2.48 \\
                & w/ C\&T & 5.68 & 22809 & 2.48 \\
            \bottomrule
        \end{tabularx}
    \end{threeparttable}
\end{table}
Based on Eq. \ref{eq:ctsn_u} - \ref{eq:g_neuromorphic} and Eq. \ref{eq:tmpr_total_loss} - \ref{eq:tmpr}, since both CTSN and TMPR do not introduce higher-order operations (e.g. $O(N^3)$), the computational complexity of CTSN and TMPR in terms of memory usage and training time are $O(LT)$, where $L$ is the number of layers and $T$ is the number of timesteps. This is identical to the baseline ternary spiking neuron and cross-entropy loss. However, the actual workload of hardware implementation varies. Since CTSN and TMPR introduce additional computations, the training time and memory usage increase accordingly. In this case, providing analysis of the computational complexity is not highly necessary, and we believe that empirical evaluation of training time and memory usage is more informative.

To evaluate the memory usage and training time of CTSN and TMPR, we conducted experiments on with 1 NVIDIA RTX 4090D GPU using ResNet20 with the experimental settings described in Section. \ref{sec:experimental_results}. The results along with the inference time on test datasets are summarized in Table. \ref{tab:appendix_computational_overhead}. Compared to baseline ternary neurons, CTSN and TMPR do introduce additional computational overhead. The training time increased by approximately 35-100\% when using CTSN and TMPR. Memory usage also increased due to the additional operations, but it remained within acceptable limits for our hardware setup. Moreover, it is worth noting that TMPR is a training method which contributes no extra computation time and memory usage during the inference stage. The inference time remains relatively unchanged, indicating that the additional computation of CTSN does not significantly impact the model's efficiency during inference. Therefore, the benefits of improved performance with CTSN and TMPR outweigh the additional computational costs, making them a valuable addition to the ternary spiking framework.

\subsection{Estimation of Energy Consumption}
\label{sec:appendix_energy_consumption}
\begin{figure}[!t]
  \centering
  \includegraphics[width=\textwidth]{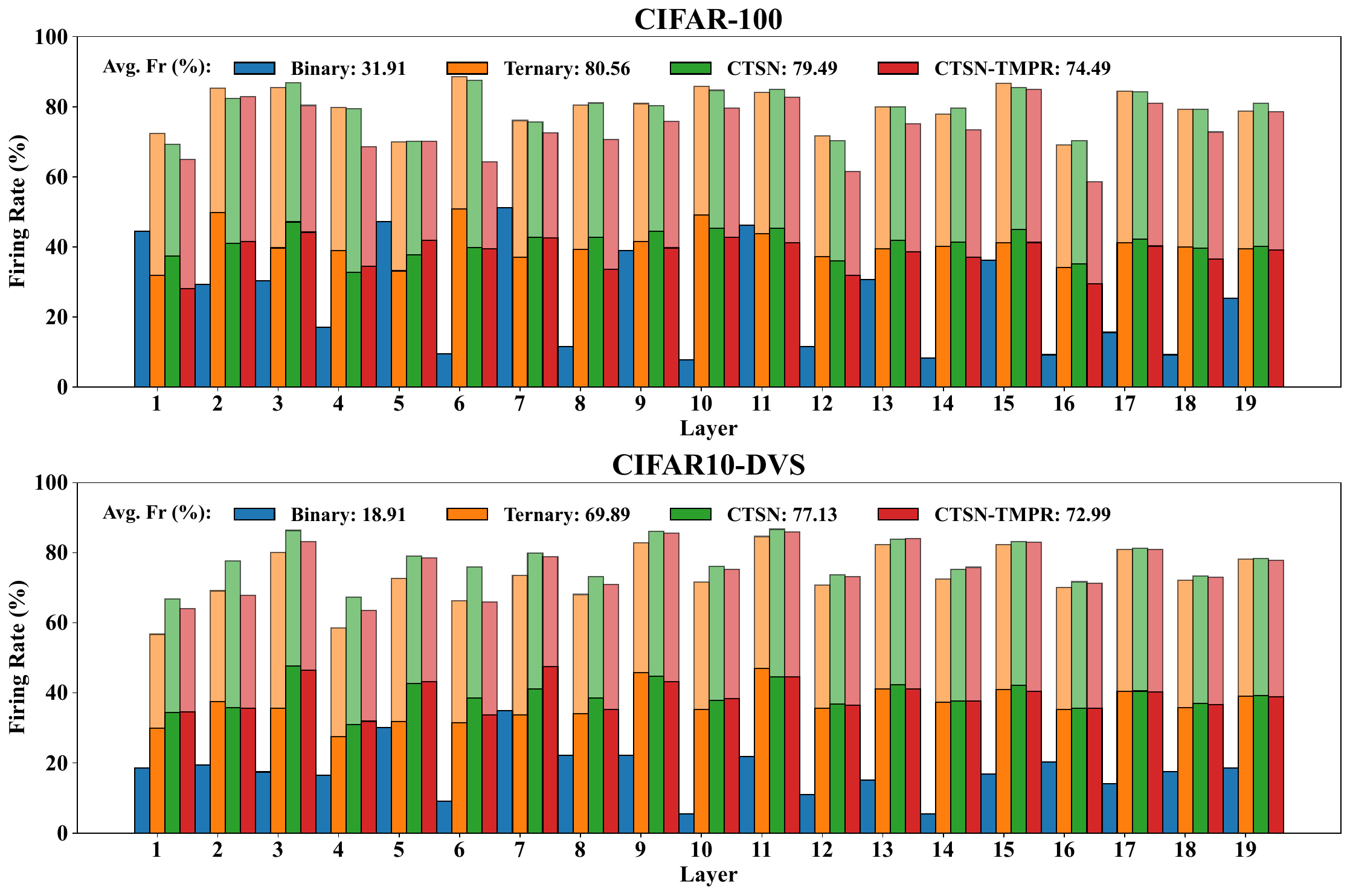}
  \captionsetup{width=\textwidth}
  \caption{Spike firing rate results. For ternary spiking neuron and our CTSN model, each bar consists of two segments: the lower segment represents +1 spike firing rate, while the upper segment represents -1 spike firing rate.}
  \label{fig:appendix_firing_rate}
\end{figure}
\textbf{Spike Firing Rate Analysis.} We conduct a quantitative comparison of the firing rates between our CTSN model and standard binary spiking neurons on both CIFAR100 and CIFAR10-DVS datasets using ResNet20. The results are summarized in Figure. \ref{fig:appendix_firing_rate}. As can be observed, the firing rate of the baseline ternary neuron is significantly higher than the binary spiking neuron, which is caused by the ternary spiking mechanism and irregularities in the membrane potential distribution. The firing rate of CTSN is closely comparable to the baseline. The addition of TMPR in training stage can reduce the firing rate.

\textbf{Energy Consumption Evaluation.} For the evaluation of energy consumption, we follow the convention of the neuromorphic computing community by counting the total synaptic operations (SOP) to estimate the computation overhead of SNN models \citep{zhou2023spikformer,huang2024clif}. In SNNs, since the input spikes are discrete, the synaptic operation is only triggered when a spike is fired. In this case, the synaptic operation is mostly accumulation (ACs) instead of multiply and accumulation (MACs). This is a significant advantage of SNNs over ANNs, as ACs are much more energy-efficient than MACs. The AC operation is defined as
\begin{equation}
  \begin{split}
    ACs &= \sum_{t=1}^{T}\sum_{l=1}^{L-1} f^l o^l(t)\\
    &= \sum_{l=1}^{L-1} f^l \cdot T \cdot \overline{o}^l(t),
  \end{split}
\end{equation}
where $f^l$ is the number of fan-out connections from the $l$-th layer to the subsequent layer, $o^l(t)$ denotes the spike output of this layer at time step $t$, and $\overline{o}^l(t)$ represents the average spike firing rate of the $l$-th layer's neuron. The MAC operation is defined as
\begin{equation}
    MACs = \sum_{l=1}^{L-1}f^l.
\end{equation}
Following the experimental setup of spike firing rate analysis, we take all the testing samples of CIFAR10-DVS and CIFAR100 to estimate the average SOP of spiking ResNet20 with binary LIF neurons, ternary spiking neurons, and CTSN neurons trained with TMPR. We assume that the ACs and MACs operations are implemented on 45nm CMOS hardware, which consumes $0.1pJ$ and $4.6pJ$ energy per operation, respectively \citep{han2015learning}. All the results are summarized in the Table. \ref{tab:appendix_energy_consumption}.
\begin{table}[!t]
  \centering
  \captionsetup{width=\textwidth}
  \caption{Estimated energy consumption. The parameters contained in the ResNet20 model are 10.8M. w/ C\&T: with CTSN \& TMPR}
  \label{tab:appendix_energy_consumption}
  \begin{threeparttable}
  \begin{tabularx}{\textwidth}{
    >{\hsize=1.3\hsize}C
    >{\hsize=0.2\hsize}C
    >{\hsize=0.8\hsize}C
    >{\hsize=0.9\hsize}C
    >{\hsize=1.1\hsize}C
    >{\hsize=1.7\hsize}C
    }
    \toprule
    \textbf{Dataset} & \textbf{T} & \textbf{Neuron} & \textbf{ACs (M)} & \textbf{MACs (M)} & \textbf{SOP Energy ($\mu$J)}\\
    \midrule
    CIFAR-100 & 4 & \makecell{LIF\\Ternary\\w/ C\&T} & \makecell{81.99\\ 216.61\\ 201.30} & \makecell{1.33\\ 1.33\\ 3.89} & \makecell{79.90\\ 201.06\\ 199.04}\\
    \midrule
    CIFAR10-DVS & 10 & \makecell{LIF\\Ternary\\w/ C\&T} & \makecell{54.83\\ 196.47\\ 204.48} & \makecell{3.20\\ 3.20\\ 9.59} & \makecell{64.07\\ 191.55\\ 228.15}\\
    \bottomrule
  \end{tabularx}
  \end{threeparttable}
\end{table}
As shown in the table, the CTSN network trained with TMPR exhibited nearly the same SOP energy as the baseline ternary spiking network on static CIFAR100. On the neuromorphic CIFAR10-DVS dataset, however, SOP energy increased owing to substantially higher AC and MAC operation counts. Overall, introducing CTSN and TMPR do not substantially compromise the energy efficiency during the inference stage. Future work should focus on further reducing energy consumption and develop ternary SNNs with efficiency comparable to that of binary SNNs.

\bibliographystyle{elsarticle-harv} 
\bibliography{references}

\end{document}